\documentclass[acmtog,screen,nonacm]{acmart}

\usepackage{booktabs} 
\usepackage{amsfonts}
\usepackage{pdfpages}

\usepackage[ruled]{algorithm2e} 

\SetAlFnt{\small}
\SetAlCapFnt{\small}
\SetAlCapNameFnt{\small}

\SetAlCapHSkip{0pt}
\usepackage{enumitem}
\setcopyright{none}
\renewcommand\footnotetextcopyrightpermission[1]{} 
\usepackage{subcaption}
\acmBooktitle{}
\begin{document}

\newcommand{\targetimg}{\mathbf{T}}  
\newcommand{\tindex}{j} 
\newcommand{\pt}{x} 
\newcommand{\viewdir}{v} 
\newcommand{\handpose}{p}  
\newcommand{\signeddist}{d}  
\newcommand{\featureop}{f_{32}} 

\newcommand{\etal}{\textit{et al.}}

\def\reta{{\textnormal{$\eta$}}}
\def\ra{{\textnormal{a}}}
\def\rb{{\textnormal{b}}}
\def\rc{{\textnormal{c}}}
\def\rd{{\textnormal{d}}}
\def\re{{\textnormal{e}}}
\def\rf{{\textnormal{f}}}
\def\rg{{\textnormal{g}}}
\def\rh{{\textnormal{h}}}
\def\ri{{\textnormal{i}}}
\def\rj{{\textnormal{j}}}
\def\rk{{\textnormal{k}}}
\def\rl{{\textnormal{l}}}
\def\rn{{\textnormal{n}}}
\def\ro{{\textnormal{o}}}
\def\rp{{\textnormal{p}}}
\def\rq{{\textnormal{q}}}
\def\rr{{\textnormal{r}}}
\def\rs{{\textnormal{s}}}
\def\rt{{\textnormal{t}}}
\def\ru{{\textnormal{u}}}
\def\rv{{\textnormal{v}}}
\def\rw{{\textnormal{w}}}
\def\rx{{\textnormal{x}}}
\def\ry{{\textnormal{y}}}
\def\rz{{\textnormal{z}}}

\def\rvepsilon{{\mathbf{\epsilon}}}
\def\rvtheta{{\mathbf{\theta}}}
\def\rva{{\mathbf{a}}}
\def\rvb{{\mathbf{b}}}
\def\rvc{{\mathbf{c}}}
\def\rvd{{\mathbf{d}}}
\def\rve{{\mathbf{e}}}
\def\rvf{{\mathbf{f}}}
\def\rvg{{\mathbf{g}}}
\def\rvh{{\mathbf{h}}}
\def\rvu{{\mathbf{i}}}
\def\rvj{{\mathbf{j}}}
\def\rvk{{\mathbf{k}}}
\def\rvl{{\mathbf{l}}}
\def\rvm{{\mathbf{m}}}
\def\rvn{{\mathbf{n}}}
\def\rvo{{\mathbf{o}}}
\def\rvp{{\mathbf{p}}}
\def\rvq{{\mathbf{q}}}
\def\rvr{{\mathbf{r}}}
\def\rvs{{\mathbf{s}}}
\def\rvt{{\mathbf{t}}}
\def\rvu{{\mathbf{u}}}
\def\rvv{{\mathbf{v}}}
\def\rvw{{\mathbf{w}}}
\def\rvx{{\mathbf{x}}}
\def\rvy{{\mathbf{y}}}
\def\rvz{{\mathbf{z}}}

\def\erva{{\textnormal{a}}}
\def\ervb{{\textnormal{b}}}
\def\ervc{{\textnormal{c}}}
\def\ervd{{\textnormal{d}}}
\def\erve{{\textnormal{e}}}
\def\ervf{{\textnormal{f}}}
\def\ervg{{\textnormal{g}}}
\def\ervh{{\textnormal{h}}}
\def\ervi{{\textnormal{i}}}
\def\ervj{{\textnormal{j}}}
\def\ervk{{\textnormal{k}}}
\def\ervl{{\textnormal{l}}}
\def\ervm{{\textnormal{m}}}
\def\ervn{{\textnormal{n}}}
\def\ervo{{\textnormal{o}}}
\def\ervp{{\textnormal{p}}}
\def\ervq{{\textnormal{q}}}
\def\ervr{{\textnormal{r}}}
\def\ervs{{\textnormal{s}}}
\def\ervt{{\textnormal{t}}}
\def\ervu{{\textnormal{u}}}
\def\ervv{{\textnormal{v}}}
\def\ervw{{\textnormal{w}}}
\def\ervx{{\textnormal{x}}}
\def\ervy{{\textnormal{y}}}
\def\ervz{{\textnormal{z}}}

\def\rmA{{\mathbf{A}}}
\def\rmB{{\mathbf{B}}}
\def\rmC{{\mathbf{C}}}
\def\rmD{{\mathbf{D}}}
\def\rmE{{\mathbf{E}}}
\def\rmF{{\mathbf{F}}}
\def\rmG{{\mathbf{G}}}
\def\rmH{{\mathbf{H}}}
\def\rmI{{\mathbf{I}}}
\def\rmJ{{\mathbf{J}}}
\def\rmK{{\mathbf{K}}}
\def\rmL{{\mathbf{L}}}
\def\rmM{{\mathbf{M}}}
\def\rmN{{\mathbf{N}}}
\def\rmO{{\mathbf{O}}}
\def\rmP{{\mathbf{P}}}
\def\rmQ{{\mathbf{Q}}}
\def\rmR{{\mathbf{R}}}
\def\rmS{{\mathbf{S}}}
\def\rmT{{\mathbf{T}}}
\def\rmU{{\mathbf{U}}}
\def\rmV{{\mathbf{V}}}
\def\rmW{{\mathbf{W}}}
\def\rmX{{\mathbf{X}}}
\def\rmY{{\mathbf{Y}}}
\def\rmZ{{\mathbf{Z}}}

\def\ermA{{\textnormal{A}}}
\def\ermB{{\textnormal{B}}}
\def\ermC{{\textnormal{C}}}
\def\ermD{{\textnormal{D}}}
\def\ermE{{\textnormal{E}}}
\def\ermF{{\textnormal{F}}}
\def\ermG{{\textnormal{G}}}
\def\ermH{{\textnormal{H}}}
\def\ermI{{\textnormal{I}}}
\def\ermJ{{\textnormal{J}}}
\def\ermK{{\textnormal{K}}}
\def\ermL{{\textnormal{L}}}
\def\ermM{{\textnormal{M}}}
\def\ermN{{\textnormal{N}}}
\def\ermO{{\textnormal{O}}}
\def\ermP{{\textnormal{P}}}
\def\ermQ{{\textnormal{Q}}}
\def\ermR{{\textnormal{R}}}
\def\ermS{{\textnormal{S}}}
\def\ermT{{\textnormal{T}}}
\def\ermU{{\textnormal{U}}}
\def\ermV{{\textnormal{V}}}
\def\ermW{{\textnormal{W}}}
\def\ermX{{\textnormal{X}}}
\def\ermY{{\textnormal{Y}}}
\def\ermZ{{\textnormal{Z}}}

\def\vzero{{\bm{0}}}
\def\vone{{\bm{1}}}
\def\vmu{{\bm{\mu}}}
\def\vtheta{{\bm{\theta}}}
\def\va{{\bm{a}}}
\def\vb{{\bm{b}}}
\def\vc{{\bm{c}}}
\def\vd{{\bm{d}}}
\def\ve{{\bm{e}}}
\def\vf{{\bm{f}}}
\def\vg{{\bm{g}}}
\def\vh{{\bm{h}}}
\def\vi{{\bm{i}}}
\def\vj{{\bm{j}}}
\def\vk{{\bm{k}}}
\def\vl{{\bm{l}}}
\def\vm{{\bm{m}}}
\def\vn{{\bm{n}}}
\def\vo{{\bm{o}}}
\def\vp{{\bm{p}}}
\def\vq{{\bm{q}}}
\def\vr{{\bm{r}}}
\def\vs{{\bm{s}}}
\def\vt{{\bm{t}}}
\def\vu{{\bm{u}}}
\def\vv{{\bm{v}}}
\def\vw{{\bm{w}}}
\def\vx{{\bm{x}}}
\def\vy{{\bm{y}}}
\def\vz{{\bm{z}}}

\title{HQ3DAvatar: High Quality Controllable 3D Head Avatar}

%%%%%%%%%%%%%%%%%%%%%%%%%%%%%%%%%%%%%%%%%%%%%%%%%%%%%%%%%%
\author{Kartik Teotia}
\affiliation{%
	\institution{Max Planck Institute for Informatics and Saarland University}
    \country{Germany}
}
\email{ktoetia@mpi-inf.mpg.de}
%%%%%%%%%%%%%%%%%%%%%%%%%%%%%%%%%%%%%%%
\author{Mallikarjun B R}
\affiliation{%
	\institution{Max Planck Institute for Informatics and Saarland University}
 \country{Germany}
}
\email{mbr@mpi-inf.mpg.de}
%%%%%%%%%%%%%%%%%%%%%%%%%%%%%%%%%%%%%%%
\author{Xingang Pan}
\affiliation{%
	\institution{Max Planck Institute for Informatics}
 \country{Germany}
}
\email{xpan@mpi-inf.mpg.de}
%%%%%%%%%%%%%%%%%%%%%%%%%%%%%%%%%%%%%%%
\author{Hyeongwoo Kim}
\affiliation{%
}
\email{}
%%%%%%%%%%%%%%%%%%%%%%%%%%%%%%%%%%%%%%%
\author{Pablo Garrido}
\affiliation{%
	\institution{Flawless AI}
\country{United States of America}
}
\email{pablo.garrido@flawlessai.com}
%%%%%%%%%%%%%%%%%%%%%%%%%%%%%%%%%%%%%%%
\author{Mohamed Elgharib}
\affiliation{%
	\institution{Max Planck Institute for Informatics}
 \country{Germany}
}
\email{elgharib@mpi-inf.mpg.de}
%%%%%%%%%%%%%%%%%%%%%%%%%%%%%%%%%%%%%%%
%%%%%%%%%%%%%%%%%%%%%%%%%%%%%%%%%%%%%%%
\author{Christian Theobalt}
\affiliation{\institution{Max Planck Institute for Informatics and Saarland University}
\country{Germany}
 }
 \email{theobalt@mpi-inf.mpg.de}
%%%%%%%%%%%%%%%%%%%%%%%%%%%%%%%%%%%%%%%

%%%%%%%%%%%%%%%%%%%%%%%%%%%%%%%%%%%%%%%%%%%%%%%%%%%%%%%%%%

\begin{teaserfigure}
\centering
  \includegraphics[width=\textwidth]{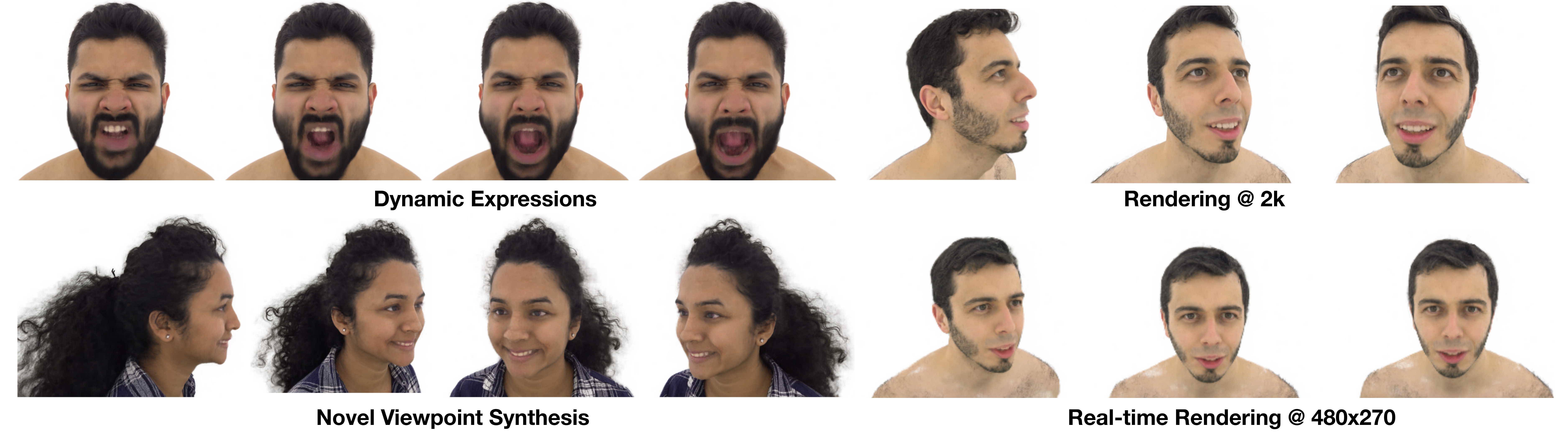}
  \caption{Our method generates a high-quality 3D head avatar that can be rendered with unseen expressions and camera viewpoints (left). It is trained using multi-view data and multiresolution hash encoding and at test it is driven by a monocular RGB video. Our approach generates 2K full-head renderings (top right) for the first time in literature. It can also run in real time at $480\times270$ (bottom right).}
  \label{fig:teaser}
\end{teaserfigure}

\begin{abstract}
   Multi-view volumetric rendering techniques have recently shown great potential in modeling and synthesizing high-quality head avatars.
A common approach to capture full head dynamic performances is to track the underlying geometry using a mesh-based template or 3D cube-based graphics primitives.
While these model-based approaches achieve promising results, they often fail to learn complex geometric details such as the mouth interior, hair, and topological changes over time. 
This paper presents a novel approach to building highly photorealistic digital head avatars. 
Our method learns a canonical space via an implicit function parameterized by a neural network.
It leverages multiresolution hash encoding in the learned feature space, allowing for high-quality, faster training and high-resolution rendering. 
At test time, our method is driven by a monocular RGB video. Here, an image encoder extracts face-specific features that also condition the learnable canonical space.
This encourages deformation-dependent texture variations during training. 
We also propose a novel optical flow based loss that ensures correspondences in the learned canonical space, thus encouraging artifact-free and temporally consistent renderings.
We show results on challenging facial expressions and show free-viewpoint renderings at interactive real-time rates for medium image resolutions.
Our method outperforms all existing approaches, both visually and numerically. We will release our multiple-identity dataset to encourage further research.

\end{abstract}

\maketitle

\section{Introduction} \label{sec:introduction}

The human face is at the center of our visual communications, and hence  its digitization is of utmost importance. Learning a high-quality controllable 3D digital head is a long-standing research problem with several applications in VR/AR, VFX, and media production, among others. Solutions to this task progressed significantly over the past few years, including early works that create a static textured face model from a monocular RGB camera~\cite{thies2016face}, all the way to recent multi-view methods that learn a highly photorealistic model, which can be rendered from an arbitrary camera viewpoint~\cite{lombardi21_mvp}.

Early methods for facial avatar creation are based on explicit scene representations, such as meshes~\cite{zollhofer18_sota-mon-3d,kim2018_dvp,thies2019_dnr}. While these methods produce photorealistic results, they cannot guarantee 3D-consistent reconstructions.
Recently, implicit scene representations have significantly attracted the attention of the research community~\cite{tewari2022_advances-neural-render}. Implicit representations assume the examined signal is continuous. Here, it was shown that, by using neural networks e.g. MLPs, high scene granularities can be reconstructed at high fidelity. Furthermore, implicit scene representations such as Neural Radiance Fields (NeRFs)~\cite{mildenhall2022_nerf} can be learned from multiple 2D images and produce multi-view consistent renderings.
These features make implicit representations suitable for the general task of 3D scene reconstruction and rendering, including human face digitization. 

Neural implicit representations~\cite{mildenhall2022_nerf,park2019_deepsdf} and, in particular, NeRF have been used for face digitization due to its high level of photorealism~\cite{gafni2021_dynamic-nerf,zheng2022imavatar,athar2022rignerf}. Here, one of the main challenges is how to model complex facial motions. Faces are dynamic objects and are often influenced by the activation of facial expressions and head poses. An early adaptation of NeRFs, applied to the human face, represents such motion by simply conditioning the implicit function, represented as an MLP, on 3DMM parameters~\cite{gafni2021_dynamic-nerf}. While this produces interesting results, it has a few limitations, primarily the inability of such 3DMMs to reconstruct high-frequency skin deformations and model the mouth interior. 
In follow-up methods, a common approach is to model motion by learning a canonical space via template-based deformation supervision~\cite{zheng2022imavatar,athar2022rignerf}. However, this kind of supervision limits the ability of these methods to accurately model regions not represented by the underlying parametric model e.g. the mouth interior.

Mixture of Volumetric Primitives (MVPs)~\cite{lombardi21_mvp} combines the advantage of mesh-based approaches with a voxel-based volumetric representation that allows for efficient rendering. Specifically, it utilizes a template-based mesh tracker to initialize voxels and prune empty spaces. Here, a primitive motion decoder modifies the initialized positions of the primitives. 
This method produces state-of-the-art results with the highest level of photorealism, mainly due to its hybrid voxel-NeRF representation as well as its capability to train on multi-view video data. 
However, finding the optimal orientation of the primitives solely based on a photometric reconstruction loss is highly challenging. As a result, this method produces inaccurate reconstructions and artifacts in regions exhibiting fine-scale details such as the hair. 
It is also expensive to train, requiring around 2.5 days when trained on an NVIDIA A40 GPU.

In this paper, we present a novel approach for producing high-quality facial avatars at state-of-the-art level of photorealism. Our approach uses a voxelized feature grid and leverages multiresolution hash encoding. It is trained using a multi-view video camera setup and, at test time, drives the avatar via a monocular RGB camera. Unlike related methods~\cite{lombardi21_mvp,gao2022_nerfblendshape}, our approach does not require a template to aid in modeling scene dynamics or pruning of empty space. Instead, we learn a fully implicit canonical space that is conditioned on features extracted from the driving monocular video. We regularize the canonical space using a novel optical flow based loss that encourages artifact-free reconstructions. Our model can be rendered under novel camera viewpoints and facial expressions during inference (see Fig.~\ref{fig:teaser}, left). It produces highly photorealistic results and outperforms state-of-the-art approaches~\cite{lombardi21_mvp,gao2022_nerfblendshape,park2021_hypernerf}, even on challenging regions such as the scalp hair. 
Our contributions are summarized as follows:
\begin{itemize}
\item We present a method that leverages a multiresolution hash table to generate volumetric head avatars with state-of-the-art photorealism. The avatar is trained using multi-view data and is driven by a monocular video sequence at test time. The core of our method is a canonical space conditioned on features extracted from the driving video. 
\item We propose a novel optical flow based loss to enforce temporal coherent correspondences in the learnable canonical space, thus encouraging artifact-free reconstructions.

\item Our model training time is 4-5 times faster than the state of the art~\cite{lombardi21_mvp}. We show a result with 2K resolution for the first time in literature. We also show a setting for rendering our results in real time (see Fig.~\ref{fig:teaser}, bottom right). 

\item We have collected a novel dataset of 16 identities performing a variety of expressions. The identities are captured using a multi-view video camera setup with 24 cameras. Our multi-view video face dataset is the first captured at 4K resolution, and we will release it to encourage further research. 
\item We show that the high level of photorealism of our model can even generate synthetic training data at high fidelity, opening the door to generalizing the image encoder to arbitrary input views for driving the avatar. 
\end{itemize}
We evaluate our approach visually and numerically against ground truth data. Here, we ablate our method with different design choices to illustrate their importance in the overall performance. Our approach outperforms existing methods~\cite{lombardi21_mvp,gao2022_nerfblendshape,park2021_hypernerf} visually and numerically, including a multi-view implementation of~\cite{,gao2022_nerfblendshape,park2021_hypernerf}.

\section{Related Work} 
\label{sec:related_work}
This section reviews prior work on photorealistic human head avatar generation, including approaches using monocular or multi-view RGB data. Early methods are based on explicit 3D scene representations, while recent ones leverage implicit representations. 

\subsection{Monocular Head Avatar Generation}
Several monocular avatar generation methods rely on explicit 3D models to estimate or regress a 3D face~\cite{thies2019_face2face,tewari2018_self-supervised,yamaguchi2018_high-fidelity,grecer2019_ganfit,tran2019_towards,shamai2019_synthesizing-facial,lattas2022_avatarme++,lin2020_towards-hf,ren2022_facial-geom} or a 3D head containing the face, ears, neck, and hair~\cite{ichim2015_dynamic_3davatar,cao2016_realtime,nagano2018_pagan} with photorealistic appearance from 2D images. These methods employ a statistical deformable shape model (a.k.a. 3DMM) of human faces~\cite{cao2014_facewarehouse,li2017_flame,gerig2018_morphable_models}
, which provides parametric information to represent the global shape and the dynamics of the face. However, explicit model-based approaches often generate avatars with coarse expressions or facial dynamics and 
usually lack a detailed representation of the scalp hair, eyes, and/or mouth interior e.g. tongue.
Other approaches attempt to synthesize dynamic full head avatars in a video via generative 2D neural rendering, driven via sparse keypoints~\cite{wang2021_one-shot,meshry2021_learned-spatial} or dense parametric mesh priors~\cite{kim2018_dvp,thies2019_dnr,tewari2020_pie,chandran2021_rendering-style}. These methods usually utilize GANs to translate parametric models into photorealistic 2D face portraits with pose-dependent appearance. Still, these methods struggle with fine-scale facial details, and they fail to generate 3D-consistent views.

Recent advances in neural implicit models for personalized head avatar creation from monocular video data have shown great promise. Most approaches learn deformation fields in a canonical space using dense mesh priors~\cite{zheng2022imavatar,athar2022rignerf,gao2022_nerfblendshape,grassal2022_neural-head}.
Here, \cite{gao2022_nerfblendshape} leverages multi-level hash tables to encode expression-specific voxel fields efficiently.
However, it still needs to regress to an intermediate expression space defined via 3DMM.
While the above methods generate photorealistic 3D heads with full parametric control, reconstructions can lack dynamics and fine-scale geometrical details, and they cannot handle extreme expressions. On the other hand, our approach is not 3DMM based and thus can model complex geometry and appearance under novel views. This is attributed to our learnable fully implicit canonical space conditioned on the driving video, as well as a novel scene flow constraint. 

\subsection{Multi-view Head Avatar Reconstruction}

A number of approaches leverage multi-view video data to create view-consistent and photorealistic human head avatars with a high level of fidelity. In the literature, we identify approaches that can reconstruct avatars from sparse views ($<=$ 10 high-resolution cameras) or require dense multi-camera systems with dozens of high-resolution views to achieve high-quality results. Due to the large volume of high-resolution video data, recent approaches have also focused on reducing computational and memory costs. Strategies such as efficient sampling~\cite{wang2021_learning-crf} and empty space pruning~\cite{lombardi21_mvp} have been proposed. We also adopt these strategies for efficient and highly detailed rendering at high resolutions.

\paragraph{\textbf{Sparse multi-view methods.}}
A line of research investigates lightweight volumetric approaches that aim at reducing the number of input views while attempting to preserve the reconstruction fidelity of dense camera approaches. Sparse methods often resort to a canonical space representation~\cite{park2021_nerfies}, which serves as a scene template for learning complex non-linear deformations. Pixel aligned volumetric avatars (PAVA)~\cite{raj21_pixel-aligned} is a multi-identity avatar model that employs local, pixel-aligned neural feature maps extracted from 3D scene locations. KeypointNeRF~\cite{mihajlovic22_keypointnerf} is another generalized volumetric avatar morphable model that encodes relative spatial 3D information via sparse 3D keypoints. At inference, both PAVA and KeypointNeRF can robustly reconstruct unseen identities performing new expressions from 2 or 3 input views. TAVA~\cite{li2022_tava} encodes non-linear deformations around a canonical pose using a linear blend skinning formulation. TAVA requires 4-10 input views to train a personalized model. While these approaches can generate photorealistic avatars with plausible dynamic deformations from sparse input views, they cannot generate fine-scale details and are sensitive to occlusions, producing renderings artifacts. We demonstrate that regions that undergo sparse sampling can still be reconstructed at high fidelity by imposing temporal coherency via optical flow.

\paragraph{\textbf{Dense multi-view methods.}} Early work with dense setups, called Deep Appearance Models (DAM) learn vertex locations and view-specific textures of personalized face models via Variational Autoencoders~\cite{lombardi2018_deep-appear}. Pixel Codec Avatars (PiCA)~\cite{ma2021_pixel-codec} improve upon DAM by decoding per-pixel renderings of the face model via an implicit neural function (SIREN) with learnable facial expression and surface positional encodings.
Most recent dense approaches adopt volumetric representations, such as discrete voxel grids~\cite{lombardi2019_neural-volumes}, hybrid volumetric models~\cite{wang2021_learning-crf,lombardi21_mvp}, or NeRFs~\cite{wang2022_morf}. Here, hybrid approaches combine coarse 3D structure-aware grids and implicit radiance functions, locally conditioned on voxel grids~\cite{wang2021_learning-crf} or template-based head tracking with differentiable volumetric raymarching~\cite{lombardi21_mvp}. In \cite{wang2022_morf}, a morphable radiance fields framework for 3D head modeling, called MoRF, is proposed. This framework learns statistical face shape and appearance variations from a small-scale database, though it demonstrates good generalization capabilities. While dense methods produce photo-realistic avatars, renderings tend to exhibit inaccuracies and blur artifacts, especially for complex structures and in infrequently observed areas, such as the scalp hair and mouth interior. Besides, most dense approaches rely on head priors, either mesh tracking or coarse voxel grids, and thus, they are prone to reconstruction errors and have limited representation power, e.g., handling details, mouth interior, and hair. Our approach overcomes existing limitations by solely relying on a well-constrained canonical representation that preserves expression semantics and scene flow correspondences.

\begin{figure*}[htb]
\centering
\includegraphics[width=\textwidth]{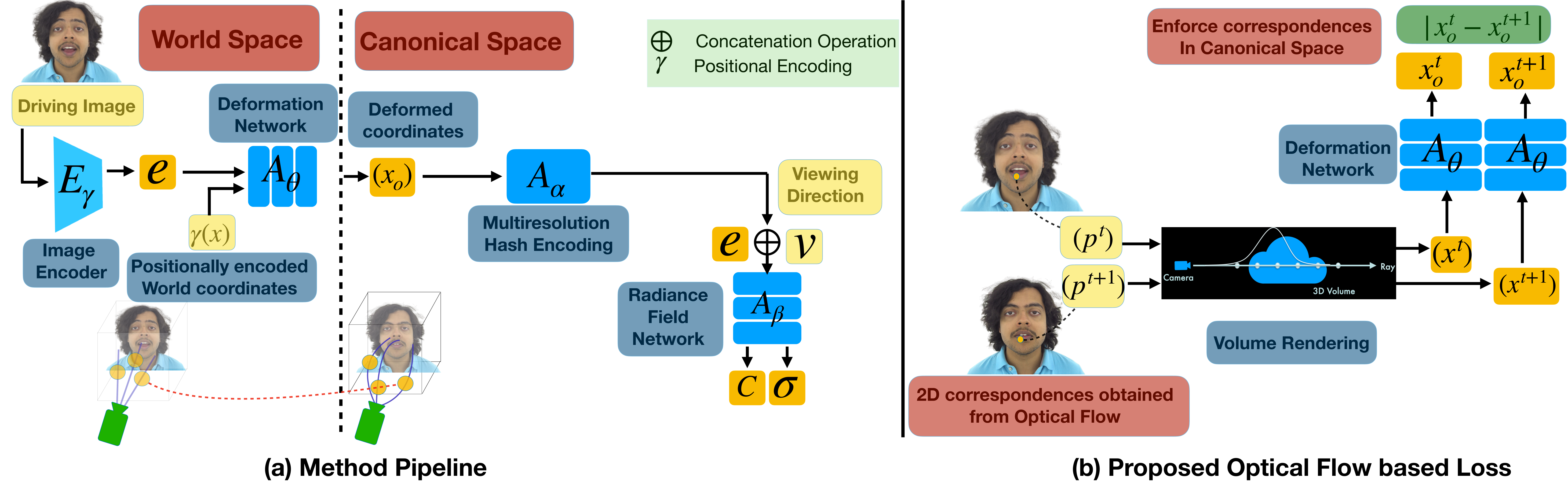}
\caption{Left: To extract a robust encoding that parameterizes the dynamics of the head, we pass a driving image through a CNN encoder to obtain a low dimensional vector \textit{e}. A deformation network $A_\theta$ conditioned on \textit{e} deforms the input coordinates $\gamma{(x)}$, where $\gamma(.)$ denotes positional encoding. 
We then use multiresolution hash encoder $A_\alpha$ to encode the deformed points in the canonical space, and feed the features from the hash grid, and encoding \textit{e} as input to a radiance field network $A_\beta$, which outputs density and color values. By combining these values through volume rendering, we are able to render the avatar under unseen input and camera viewpoints. Right: We impose a novel scene flow based constraint by utilizing the optical flow at frame $t$ and $t+1$ (see Eq.~\ref{eq:flow}). Such constraints enforce good correspondences in the canonical space, thus reducing rendering artifacts.} 

\label{fig:method}
\end{figure*}

\subsection{Generalized 3D Consistent Neural Representations}
Modeling 3D-aware scenes with implicit models has been active research in recent years. Popular methods are NeRFs~\cite{mildenhall2022_nerf} and neural Signed Distance Functions (SDFs)~\cite{park2019_deepsdf}, both parameterize the 3D space using multi-layer perceptrons (MLPs). Since such methods are often computationally expensive, efficient feature and/or scene space encodings, such as hash grids~\cite{mueller2022_instant-ngp,fridovich2022_plenoxels} or trees~\cite{yu2021_plenoctrees,takikawa2021_nglod}, have been proposed to boost performance. 
In the literature, generalized implicit models for head avatar reconstruction are learned from a large corpus of 2D face images with varying pose and facial shape using neural SDFs~\cite{ramon2021_h3dnet, or-el2022_stylesdf}, GAN-based NeRFs~\cite{deng2022_gram, chan2021_pigan, gu2022_stylenerf} or hybrid volumetric approaches with tensor representations~\cite{chan2022_efficient-geom-aware,wang2021_learning-crf}. Generalized models often lack personalized details. However, they have proven themselves to be robust priors for downstream tasks, such as landmark detection~\cite{zhang2022_flnerf}, personalized face reenactment~\cite{bai2022_high-fidelity} and 3D face modeling~\cite{abdal2023_3davatargan}. 

We remark that NeRFs have stood out as superior implicit representations for head avatar creation as they excel at reconstructing complex scene structures. Some recent prior-free NeRF-based methods focus on generating detailed avatars from very sparse 2D imagery, e.g., using local pixel-aligned encodings~\cite{raj21_pixel-aligned, mihajlovic22_keypointnerf}, while others model dynamic deformations when working with unstructured 2D videos by warping observed points into a canonical frame configuration~\cite{park2021_nerfies,park2021_hypernerf} or modeling  time-dependent latent codes~\cite{li2021_neural-scene,li2022_neural-3d-video}. We remark that dynamic approaches, while achieving impressive results, are designed to memorize the scene representations and cannot control the model beyond interpolations.
In addition, some approaches build upon dynamic NeRF approaches by incorporating parametric models, e.g., 3DMMs~\cite{egger2020_3dmfm,li2017_flame}, as input priors to enable full facial control~\cite{hong2021_headnerf,sun2022_controllable}.

\section{Method} \label{sec:method}

Let $\{I_j^i\}$ $(j=1 \ldots N, i=1 \ldots M)$ be multi-view frames of a person's head performing diverse expressions, where $N$ is the number of frames and $M$ is the total number of cameras. Our goal is to create a high-quality volumetric avatar of the person's head, which can be built in a reasonable time and rendered under novel views and expressions at unprecedented photorealism and accuracy.
Humans are capable of performing extremely diverse and extreme expressions. Our model should be able to capture these in a multi-view consistent manner with a high degree of photorealism.
As shown in Fig.~\ref{fig:method} (a), we have 4 components. Our model drives the avatar from a monocular image encoded via a CNN-based image network $E_\gamma$. We then have an MLP-based deformation network $A_\theta$, which can map a point in the world coordinate system to a canonical space conditioned on the image encoding. We learn features in the canonical space using a multiresolution hash grid $A_\alpha$. The features in the grid are interpreted to infer color and density values using an MLP-based network $A_\beta$. 
Given any camera parameters, 
we use volumetric integration to render the avatar.
In the following, we provide details about the capture setup and data pre-processing step (Sec.~\ref{subsec:dataset}), describe the scene representation of our model (Sec.~\ref{subsec:representation}), and formulate various objective functions used for model training (Sec.~\ref{subsec:obj_functions}).

\begin{figure*}[htb]
\centering
\includegraphics[width=\textwidth]{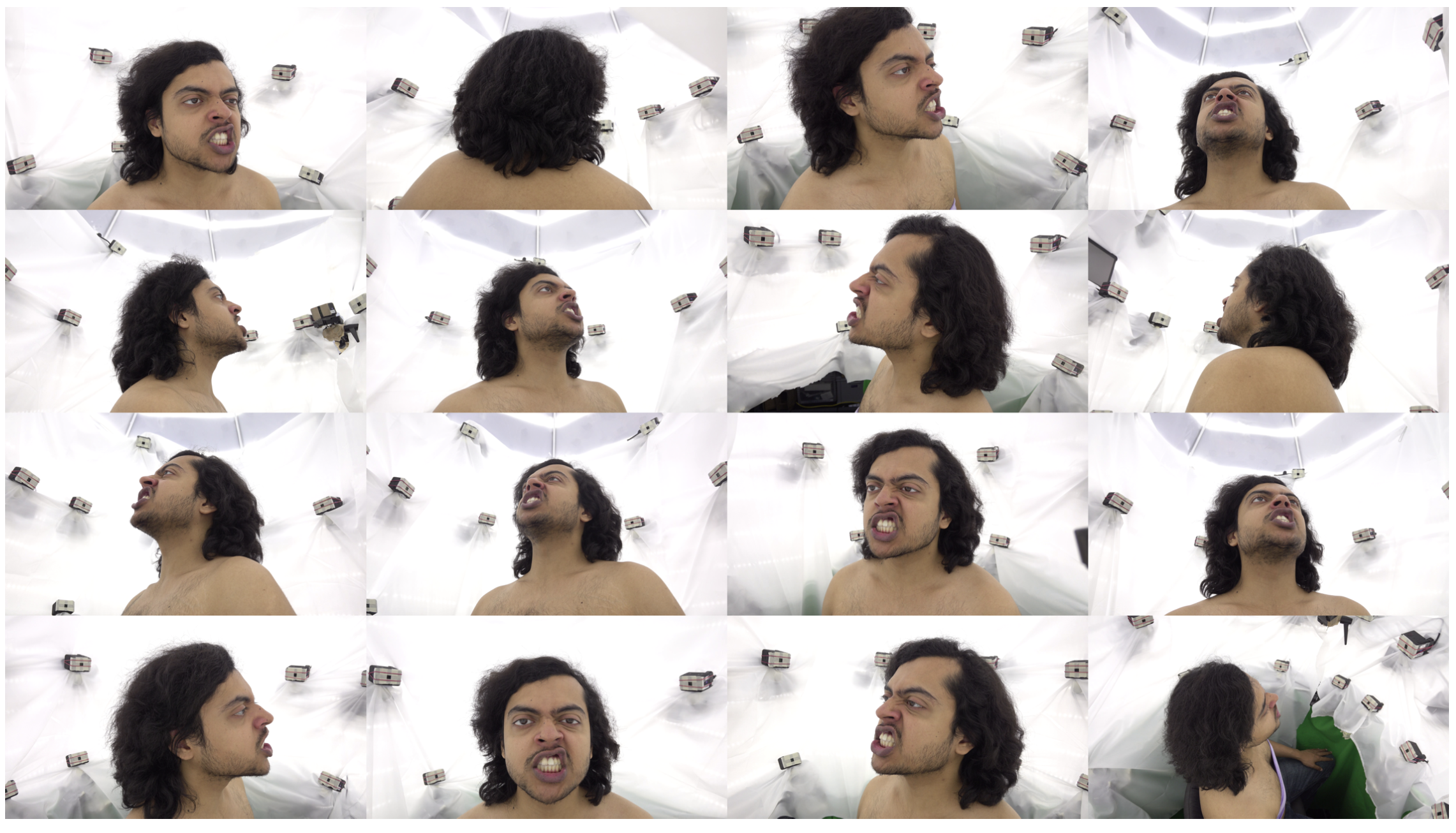}
\caption{An example of our camera rig capturing the same expression from 16 different viewpoints.}
\label{fig:rig}
\end{figure*}
\subsection{Data Capture}
\label{subsec:dataset}
\paragraph{\textbf{Capture Setting.}}
Our approach is trained using multi-view images captured from a 360-degree camera rig.
The rig is equipped with 24 Sony RXO II cameras, which are hardware-synced and record 4K resolution videos at $25$ frames per second. The cameras are positioned in such a way that they capture the entire human head, including the scalp hair. The rig is covered by LED strips to ensure uniform illumination. In our setup, we recorded a total of $16$ identities performing a wide variety of facial expressions and head movements.
Please see Fig.~\ref{fig:rig} for a sample identity captured from multiple viewpoints. For a more detailed description of our dataset, please refer to Sec.~\ref{sec:datasets}.

\paragraph{\textbf{Preprocessing}.}
Cameras are calibrated using a static structure with a large number of distinctive features. Here, we use Metashape \cite{metashape} to estimate the extrinsic and intrinsic parameters. We also perform background subtraction using the matting approach of Lin~\etal~\cite{lin2021_back_mattv2} to remove any static elements from the scene, e.g., wires, cameras, etc. To simplify background subtraction, a diffused white sheet was placed inside the rig, with holes for each of the camera lenses.

\subsection{Scene Representation}
\label{subsec:representation}

We parameterize our model using Neural Radiance Fields inspired by the state-of-the-art novel view synthesis method NeRF~\cite{mildenhall2022_nerf}. Since the original method is slow to train and render, we utilize a multiresolution hash grid-based representation to make our model efficient, akin to instant NGP \cite{mueller2022_instant-ngp}.
As both original NeRF and instant NGP were proposed for static scene reconstruction, we seek to model the dynamic performance of the head, including facial expressions. 
To this end, we represent our model, $A$ as 
\begin{equation}
    A : (\rvx, \rvv, \rve) \rightarrow (\rvc, \sigma) \; \text{,}
\end{equation}
where $\rvx \in \mathbb{R}^3$ is a point in $3D$, $\rvv \in \mathbb{S}^2$ is the viewing direction,
$e \in \mathbb{R}^{256}$ represents the latent vector obtained from the image encoding network $E_\gamma$. This latent vector parameterizes deformations due to expressions and head movements.
Furthermore, $\rvc$ and $\sigma$ are the color and density values, respectively.  
Mathematically, instant NGP parameterizes $A$ with two modules. The first module is based on a multiresolution hash grid, denoted $A_\alpha$, and the second module is parameterized by an MLP, denoted $A_\beta$. The latter takes features looked up from $A_\alpha$ and decodes a given point $\rvx$ and view direction $\rvv$ into $\rvc$ and $\sigma$.
To model dynamic variations of the input driving performance, we introduce another module, denoted $A_\theta$, which takes as input a point in world space and expression latent vector, and regresses a deformation field that converts the world point $x$ to a canonical space, as follows:
\begin{equation}
\label{eq:can_point}
    \rvx_o = A_\theta(\rvx, \rve) + \rvx \; \text{.}
\end{equation}
We learn the radiance field in this canonical space using $A_\alpha$ and $A_\beta$, and parameterize the operator $A_\theta$ using a linear MLP.
One could also naively provide the driving image latent code directly to $A_\beta$ instead of modeling a deformation field to canonical space. 
However, we show in our experiments (see Sec.~\ref{sec:ablation_study}) that such a naive parameterization creates artifacts. Thus, learning a deformation field is critical in reducing the artifacts.

Once we have the radiance field representation of the scene, we use standard volumetric integration to synthesize color $\rmC$  for each ray $\rvr(t) = \rvo + t\rvd$, with near and far bounds $t_n$ and $t_f$, as follows:
\begin{align}
\rmC(\rvr) &= \int_{t_n}^{t_f} T(t) \sigma(\rvr(t)) \rvc(\rvr(t)) dt \nonumber \; \text{,}\\
\text{where} \quad T(t) &= \text{exp}(-\int_{t_n}^{t}\sigma(\rvr(s)) ds).
\label{eq:nerf_vol_int}
\end{align}

\paragraph{\textbf{Efficient ray marching.}}
As in instant NGP, we improve efficiency by skipping regions that do not contribute to the final color based on the coarse occupancy grid. The occupancy grid typically spans $64^3$ resolution, with each cell represented by a single bit. 
The occupancy grid is updated at regular intervals by evaluating the density of the model in the corresponding region in space.
The high in each bit represents the corresponding $3D$ region that has density above a certain threshold. Note that only these regions contribute to the final rendering.  
As our scene is dynamic, we make certain changes to suit this setting. We initialize $G$ separated occupancy grids corresponding to $G$ uniformly sampled frames. 
We update each of these grids independently for $200,000$ iterations. 
Then, we take the union of all the grids to create a single occupancy grid 
 that we utilize for the rest of the training and novel view synthesis.

\subsection{Encoder}
\label{subsec:encoder}
Our model is conditioned on a latent vector $\rve$ to drive the avatar.
In the literature, some methods use expression parameters obtained from face tracking using an existing morphable model~\cite{gafni2021_dynamic-nerf, athar2022rignerf}.
Other methods parameterize the latent vector obtained from an image encoder~\cite{raj21_pixel-aligned}. 
Using an image encoder is advantageous since it can capture diverse expressions as opposed to expression parameters obtained from a 3DMM. 
Typically, tracking pipelines utilize linear morphable models that have limited expressivity and are prone to tracking errors \cite{mallikarjun2021_learning-complete}.
In this paper, we rely on image encoder $E_\gamma$ to parameterize the dynamics of the human head 
because it allows us to capture diverse and extreme expressions faithfully, which is the main focus of our paper. We parameterize $E_\gamma$ using a CNN-based network, which receives as input an image $\rmI$ and outputs the encoding vector $\rve$. 
Specifically, we adopt a pre-trained VGG-Face model~\cite{Parkhi15} as our encoder and add a custom linear layer at the end. During training, we finetune all the VGG layers as well as the custom layer.

\subsection{Objective Function}
\label{subsec:obj_functions}

Given the above representation of our model, we learn the parameters of $E_\gamma, A_\theta, A_\alpha$, and $A_\beta$ modules in a supervised manner using multi-view image and perceptual constraints as well as dense temporal correspondences:

\begin{equation}
    \mathcal{L} = \mathcal{L}_{L2} + \lambda_{perc} \mathcal{L}_{perc} + \lambda_{of} \mathcal{L}_{of} \; \text{.}
    \label{eq:obj_function}
\end{equation}

\paragraph{\textbf{Reconstruction Losses.}} 
Given camera extrinsic and model representation, we render images and employ image reconstruction loss, $\mathcal{L}_{L2}$ using L2 loss between ground truth and rendered images. This term introduces multi-view constraints to train our model. 
However, L2 loss alone could result in missing some high-frequency details, which are perceptually very important.
As a result, we introduce a widely used patch-based perceptual loss $\mathcal{L}_{perc}$, based on a pre-trained VGG Face network~\cite{Parkhi15}. We use the output of the first $6$ layers obtained from an input patch size of $64\times64$ to compute this loss term.

\paragraph{\textbf{Optical flow based Loss.}}
As our dataset consists of sparse views and hash grid-based representation has localized features, 
a model trained only with $\mathcal{L}_{L2}$ and $\mathcal{L}_{perc}$ losses tend to overfit training views, resulting in artifacts when rendering novel views.
To mitigate it, we propose a novel loss term $\mathcal{L}_{of}$ based on pre-computed $2D$ optical flow between concurrent frames. 
The motivation behind this loss term is to propagate pixel correspondences to the 3D canonical space with the aim to regularize the dynamic scene and mitigate the model's artifacts when trained with sparser views.
We achieve this by enforcing the canonical points of neighboring temporal frames to be close to each other for the points near the surface of the avatar.

Mathematically, let $p^t$, $p^{t+1}$ be the corresponding pixels between consecutive frames obtained using $2D$ optical flow. For these pixels, we first obtain their corresponding expected depth values through volume rendering. The corresponding $3D$ points $x^t$, $x^{t+1}$ associated with expected depth can be considered to be close to the surface. We find the corresponding points in the canonical space using $A_\theta$, as defined in Eq.~\ref{eq:can_point}. Let $x_o^t$ and $x_o^{t+1}$ be the corresponding points in the canonical space. We enforce all such points to be close between them by employing an L1 loss, similar to ~\cite{kasten2021layered}:
\begin{equation}
\label{eq:flow}
    \mathcal{L}_{of} = \| x_o^t - x_o^{t+1} \|_1 \; .
\end{equation}

Please refer to Fig.~\ref{fig:method} (b) for an illustration of the proposed loss term.

\subsection{Implementation Details}\label{sec:implementation_details}

We use $5$ layered MLP as our deformation network $A_\theta$.
We provide hash encoding parameters and their ranges used in our experiments in Tab.~\ref{tab:hashgrid_parameters}.
Our radiance field network $A_\beta$ is parameterized by a $5$ layer-deep MLP.
We set $\lambda_{perc}=0.1$ and $\lambda_{of}=0.2$ in our experiments.
We also follow a PyTorch implementation~\cite{tang2022_torch-ngp} of instant NGP~\cite{mueller2022_instant-ngp} to employ error map-based pixel sampling while training, for better convergence. Specifically, we maintain a $128$x$128$ resolution error map for each training image, which is updated in every iteration to reflect the pixel-wise $L_2$ error. This is then used to sample rays where errors are the highest at each iteration. 
Finally, we update our encoder $E_\gamma$, deformation network $A_\theta$, hash grid $A_\alpha$ and radiance field $A_\beta$ with learning rates $1e-5$, $1e-3$, $1e-2$ and $1e-3$, respectively. Our model is trained for $500,000$ iterations.
We have observed that model convergence is faster than in MVP~\cite{lombardi21_mvp}. It takes about $12$ hours to converge, as opposed to the $50$ hours required by MVP with the same GPU resources.

\begin{table}[]
\centering
\begin{tabular}{ll}
\toprule
    Parameter                                   &   Values \\ \hline
    Number of levels                            &   $16$     \\
    Max. entries per level (hash table size)    &   $2^{14}$   \\
    Number of feature dimensions per entry      &   $2$     \\
    Coarsest resolution                         &   $16$     \\      
    Finest resolution                           &   $2048$     \\ \hline
\end{tabular}
\caption{Different parameters used for defining the hash grid.}
\label{tab:hashgrid_parameters}
\end{table}

\section{Experiments} \label{sec:experiments}

\begin{figure*}[!ht]
\centering
\includegraphics[width=\textwidth]{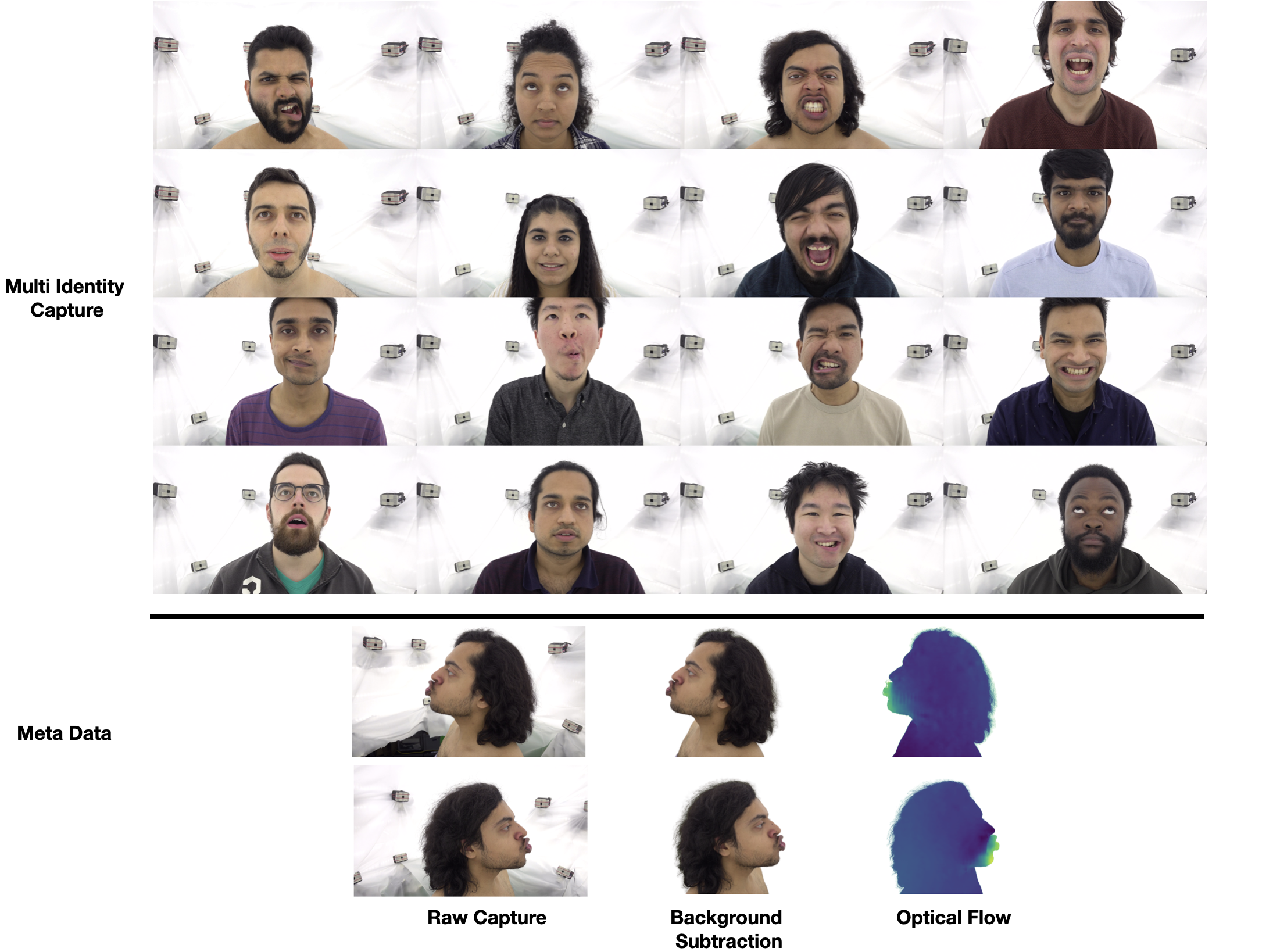}
\caption{Top: Visualization of all identities captured in our multi-view camera setup. Our dataset captures a variety of facial hair, scalp hair, expressions, and ethnicities, among others. Bottom: Example of meta data released with our dataset.}
\label{fig:datasetIDs}
\end{figure*}

In this section, we show the effectiveness of our high-quality volumetric head avatar reconstruction method in synthesizing novel dynamic expressions and views at high fidelity and resolution. We show two main applications our approach enables, namely dynamic free-view synthesis from arbitrary monocular viewpoints as well as renderings at different image resolutions, including FHD. We also perform a thorough analysis of our modeling choices and conduct quantitative and qualitative evaluations with state-of-the-art baselines. We refer the reader to the supplemental for video results.

\subsection{Datasets}\label{sec:datasets}
Our multi-view video dataset consists of $16$ subjects, including $14$ males and $2$ females, and most of them are in their 20s or 30s. 
The subjects have short to long-length hairstyles. Male subjects either are shaved or have stubble or hairy beards. A collage of the recorded subjects is shown in Fig.~\ref{fig:datasetIDs}, top. 
To build our dynamic dataset, we instructed subjects to perform random expressive faces during $2$ minutes and/or recite $47$ phonetically balanced sentences. 
Among the 16 subjects, 4 have only performed expressions, 1 has only performed reciting, while 11 have performed both. We will release our full multi-view video dataset to foster future research on head avatar generation. For all of our experiments reported next, we utilize $18$ views, each containing $1500$ frames at 960x540 resolution, to train our personalized models and generate results, unless stated otherwise. We processed 6 subjects covering a wide variety of our dataset e.g. gender, expressions, facial hair, movements, scalp hair, ethnicity, etc. 

\subsection{Qualitative and Quantitative Results}\label{sec:qualitative_evaluations}

\begin{table}[!ht]
\centering
\begin{tabular}{|c|c|c|c|c|}
\hline
Metrics & Without  & Without & Without   & Ours   \\
 & canonical &  image feature & optical flow   &    \\
  & space &conditioning & based loss &    \\
\hline
PSNR $\uparrow$  &29.24 & 29.64 &29.38 &\textbf{31.23} \\
\hline
L1 $\downarrow$ &3.61& 3.64& 3.32&  \textbf{2.79}  \\
\hline
SSIM $\uparrow$ &0.8698& 0.8744& 0.8517& \textbf{0.8837}  \\
\hline
LPIPS $\downarrow$ &0.1408& 0.1191&0.1200 & \textbf{0.1130}   \\
\hline
\end{tabular}
\caption{Ablation study: Image quality and perceptual metrics for different design choices. L1 measures the absolute error of unnormalized RGB images. Our full method produces the best results (see bold text).}
\label{tab:quantitative_results-ave_error-ablation}
\end{table}

\begin{figure*}[!ht]
\centering
\includegraphics[width=\textwidth]{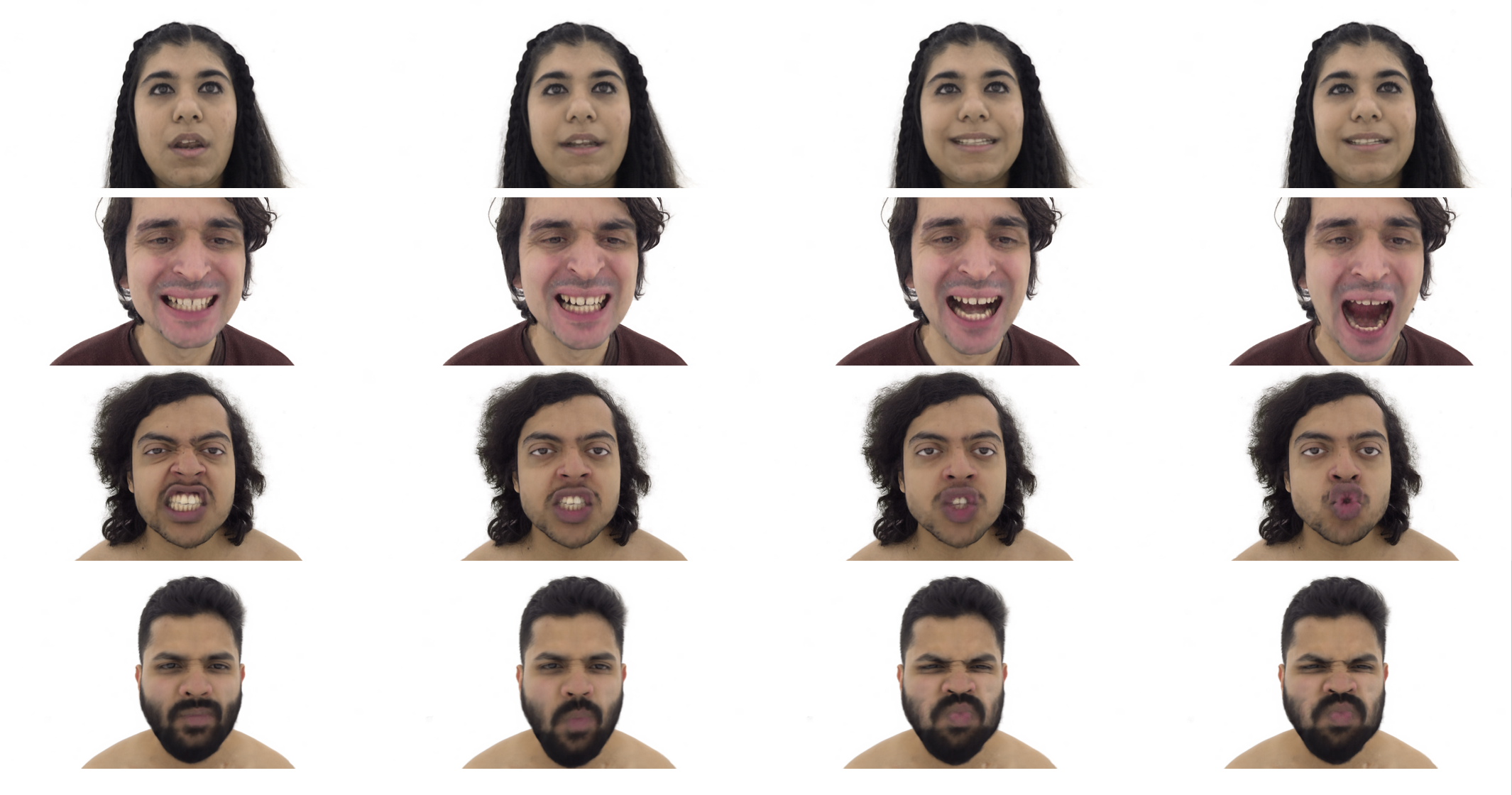}
\caption{Qualitative results: Dynamic expression changes. \textit{Bottom to top}: \textit{Subject} $1$, $2$, $3$, and $4$.}
\label{fig:qualitative_results-exp_changes}
\end{figure*}

\begin{figure*}[!ht]
\centering
\includegraphics[width=\textwidth]{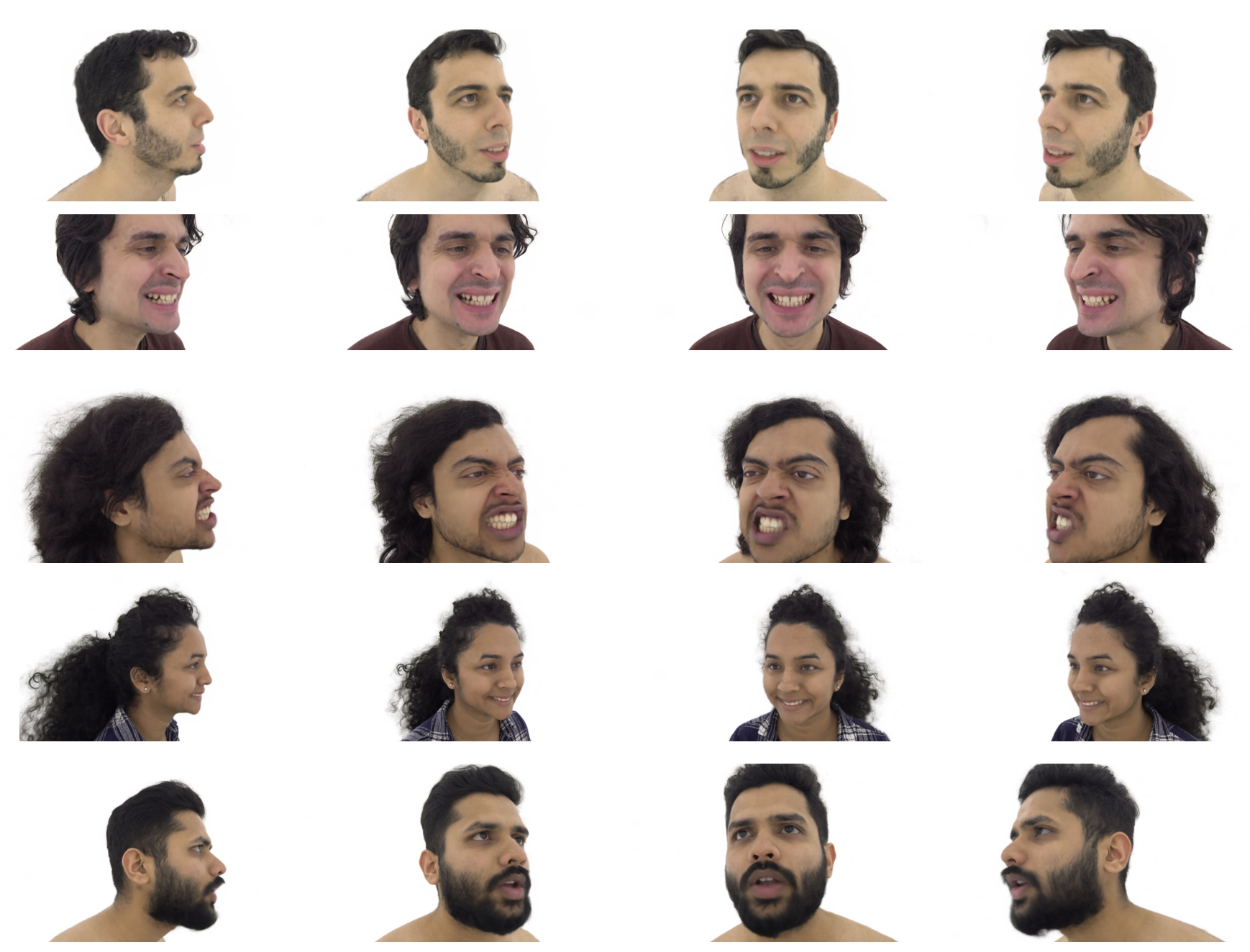}
\caption{Qualitative results: Dynamic novel view synthesis for different subjects. \textit{Bottom to top}: \textit{Subject} $5$, $2$, $3$, $6$, and $4$.}
\label{fig:qualitative_results-nvs}
\end{figure*}

Fig.~\ref{fig:qualitative_results-exp_changes} shows dynamic expression synthesis of $4$ personalized avatar models on test sequences, while Fig.~\ref{fig:qualitative_results-nvs} illustrates free viewpoint synthesis of $5$ personalized models. Note that the generated views represent interpolations from training views. In both figures, the avatars are driven by a frontal-looking monocular RGB video. Our approach achieves high-quality renderings of head avatars under novel camera viewpoints and for challenging novel expressions. Tab.~\ref{tab:quantitative_results-ave_error-ablation} shows that our approach on average obtains high PSNR (over $31$ dB) and low reconstruction errors on test sequences based on different image quality and perceptual metrics. Please see the supplemental for video results.

\subsection{Applications}\label{sec:applications}
\paragraph{\textbf{Avatar Synthesis from an Arbitrary Monocular Viewpoint}.}

In previous experiments, we have shown that we can drive our head avatar using a monocular video captured from a frontal view.
Here we further show an application where we can drive our head avatar from an arbitrary viewpoint.
To achieve this, we define a fine-tuning scheme described as follows: 
First, we synthesize a training dataset from a novel viewpoint, say $\hat{v}$, with the personalized avatar model described in Sec.\ref{sec:method}. This dataset contains the same dynamic expressions used for training. Then, we finetune the image encoder with this synthetic video stream for 100k iterations. Note that the deformation and radiance field networks as well as the multiresolution hash encoding remain unchanged.
Once the image encoder has been fine-tuned, we can drive the personalized avatar model with the real data stream coming from the viewpoint $\hat{v}$. 
In our experiments, $\hat{v}$ is a held-out viewpoint not used when training the avatar model. 

\begin{figure*}[htb]
\centering
\includegraphics[width=\textwidth]{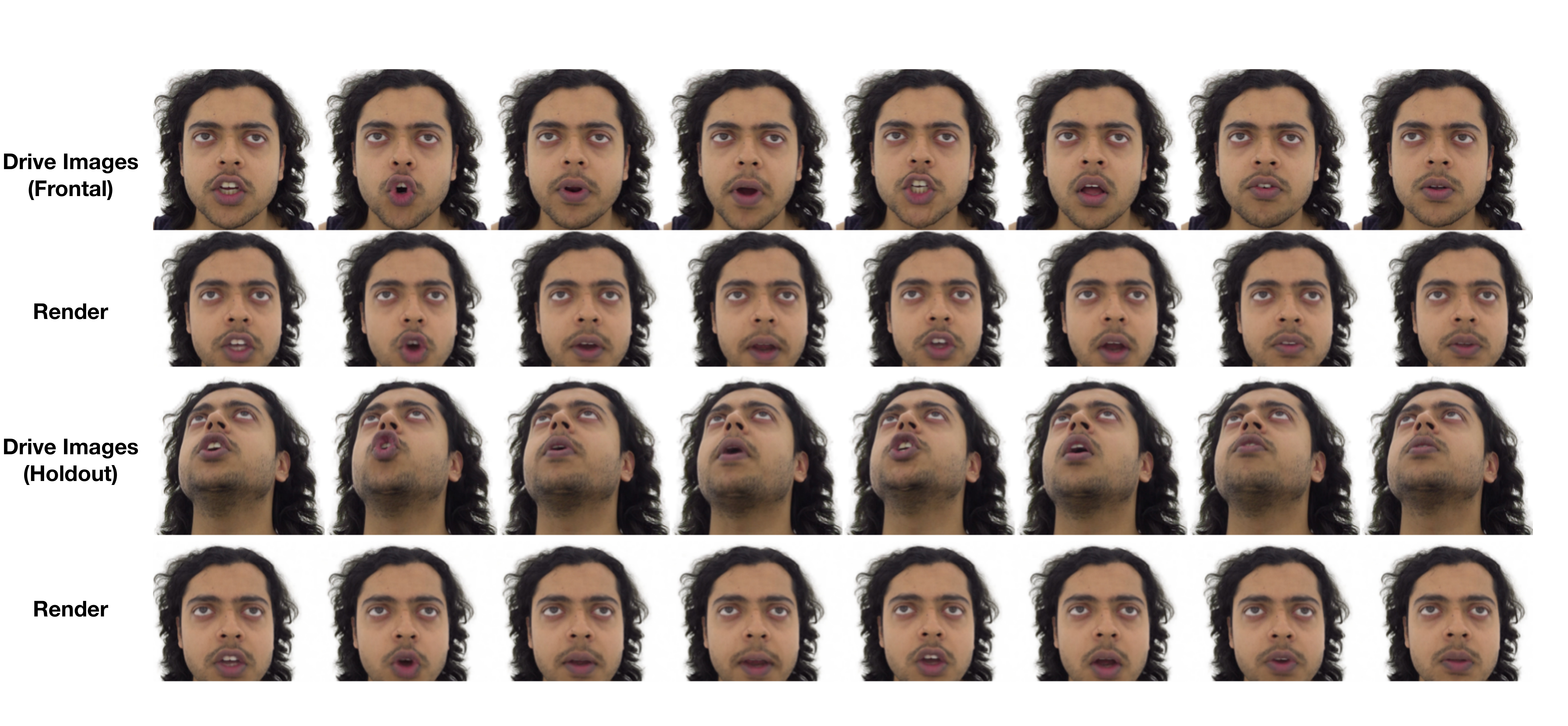}
\caption{Avatar synthesis from different driving viewpoints. \textit{Top}: Frontal view driving video and frontal rendering. \textit{Bottom}: Bottom view driving video and  frontal rendering.}
\label{fig:qualitative_results-drive_finetuned}
\end{figure*}

Fig. \ref{fig:qualitative_results-drive_finetuned} compares frontal renderings of Subject 3's avatar model, driven from two video streams with unseen expressions: One driven from a frontal view camera and another driven from a held-out bottom view. Our method produces high-fidelity renderings regardless of the driving video viewpoint, and the rendered expressions faithfully reproduce those shown in the driving video. 
This demonstrates that our personalized avatar model can generate photo-realistic renderings from arbitrary viewpoints at high fidelity. These renderings can be used as a good approximation of real images to fine-tune the image encoder from arbitrary driving viewpoints. Note that this experiment paves the way for learning high-fidelity personalized avatars that can be driven from video captured in the wild. 

\paragraph{\textbf{FHD Image Synthesis.}}

\begin{figure*}[htb]
\centering
\includegraphics[width=\textwidth]{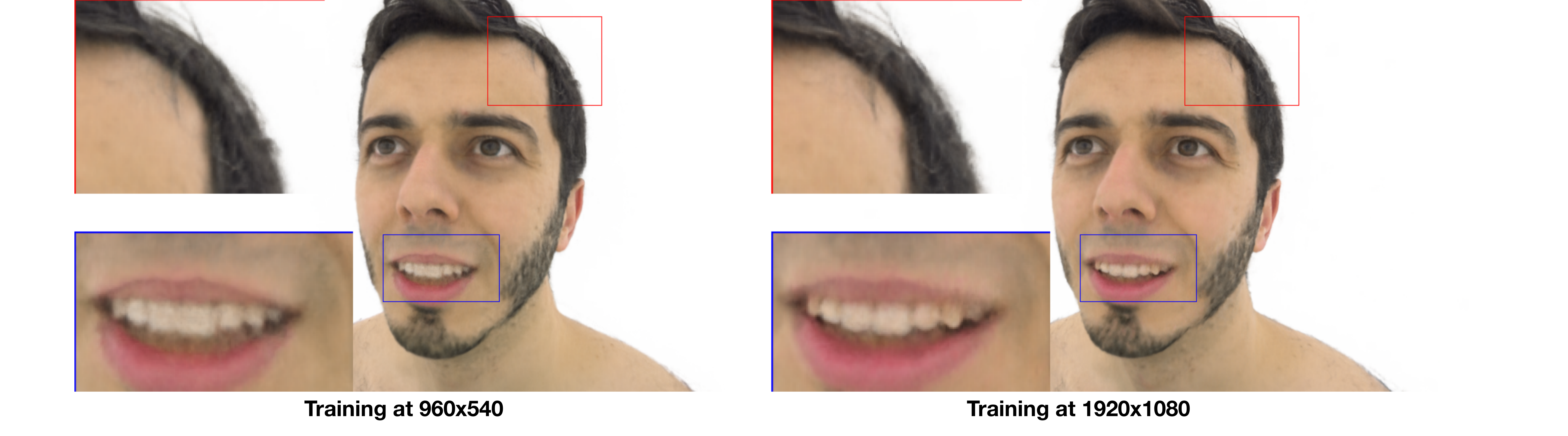}
\caption{Avatar synthesis at different resolutions. \textit{Left to right}: Model trained at HD and FHD resolutions, respectively.}
\label{fig:qualitative_results-FHD_training}
\end{figure*}

\begin{figure*}[htb]
\centering
\includegraphics[width=\textwidth]{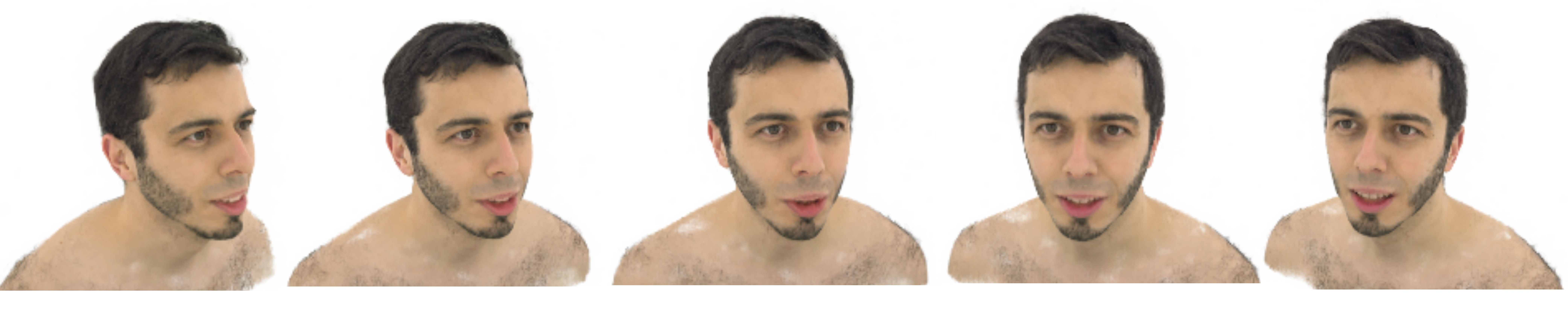}
\caption{Real-time rendering ($25$ fps) at 480x270 resolution}
\label{fig:realtime}
\end{figure*}

Our multiresolution hash grid encoding allows for training a personalized avatar model at full HD resolutions, which surpasses the capabilities of state-of-the-art approaches. Our method can render HD images (960x540) at about $10$ fps and FHD (1920x1080) images a bit below $3$ fps. Fig.~\ref{fig:qualitative_results-FHD_training} compares renderings of personalized models trained at HD and FHD resolutions. Both models generate visually similar facial features and details, though the FHD model produces crisper results, as expected. Overall our approach scales well, and the decrease in runtime is near linear. Fig.~\ref{fig:realtime} shows that our approach can also run on a resolution of 480x270 in real time ($25$ fps) while still maintaining high fidelity in the reconstructions. Note that the reported runtimes are based on a single NVIDIA A100 GPU. Please see the supplemental video for more results.

\subsection{Ablative Analysis} 
\label{sec:ablation_study}

\begin{figure*}[htb]
\centering
\includegraphics[width=\textwidth]{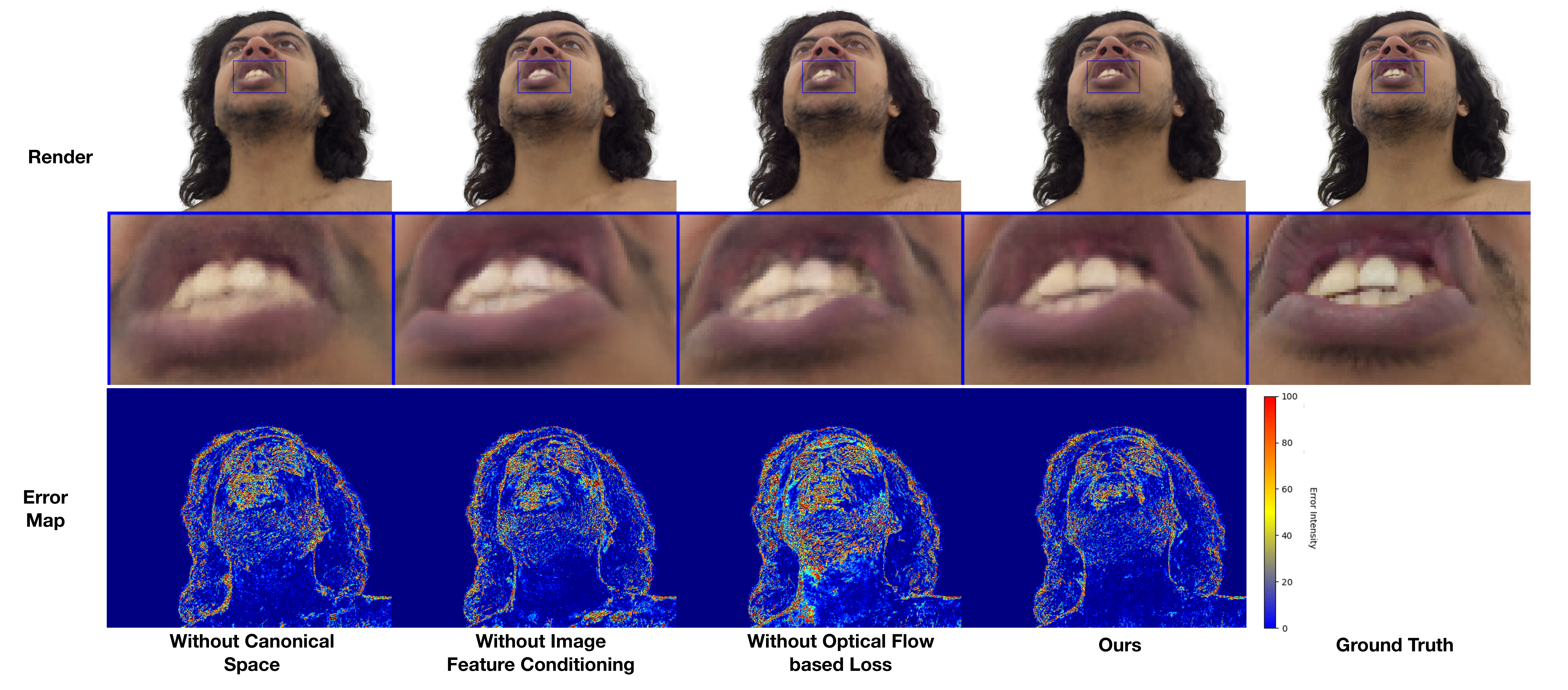}

\caption{Ablation study: Fixed view image synthesis for different design choices. \textit{Left to right}: Without canonical space, without feature conditioning, without optical flow based loss, and ours. The top row shows a rendering of Subject 3 (and ground truth), while the bottom row shows the error map. The error is computed as the per-pixel mean squared error (MSE), encoded in RGB color space. Here, blue denotes $0$ MSE, yellow is $60$ MSE, and reddish colors mean over $100$ MSE. Our full method achieves the best results.}
\label{fig:ablation_study-fvs_error-canonical_feature_oflow_ours}
\end{figure*}

We demonstrate our main contributions and the influence of design choices via a number of ablation studies. Specifically, we study our novel optical flow based loss, learned image based feature conditioning of the canonical radiance field network, and canonical space representation. We also analyze the influence of perceptual loss and error map based pixel sampling in the reconstruction quality. Note that for these experiments, we train our personalized avatar models on $18$ views, while we keep out $2$ views for our quantitative evaluations.

\begin{figure*}[htb]
\centering
\includegraphics[width=\textwidth]{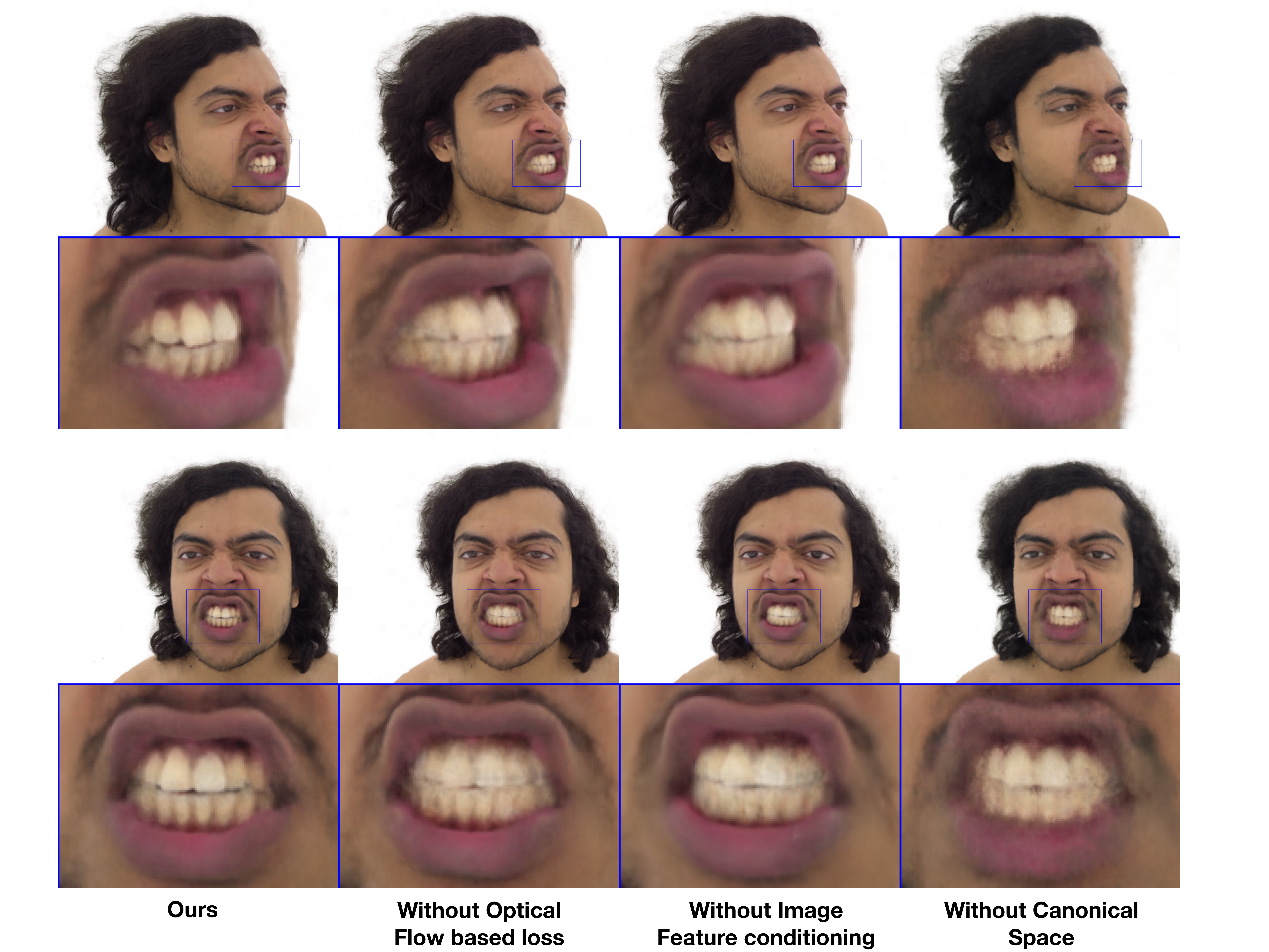}
\caption{Ablation study: Novel view synthesis quality. \textit{Left to right}: Ours, without optical flow based loss, without image feature conditioning, and without canonical space. Our full method achieves the best results.}
\label{fig:ablation_study-dvs-perceptual_error_ours}
\end{figure*}

\begin{figure*}[htb]
\centering
\includegraphics[width=0.9\textwidth]{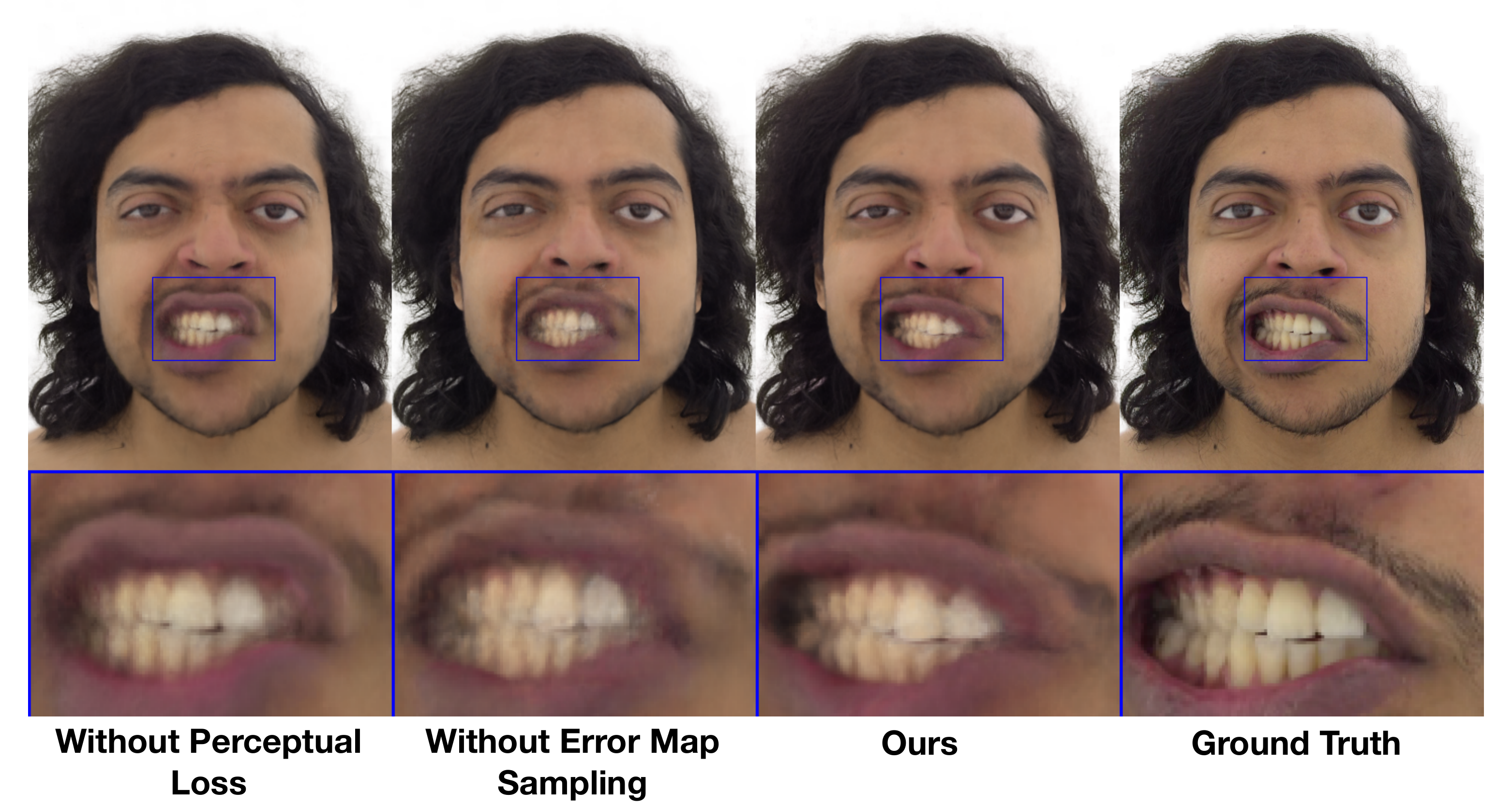}
\caption{Ablation study: Structural consistency and detail quality. \textit{Left to right}: No perceptual loss, no error map sampling, ours, and ground truth. }
\label{fig:ablation_study-details-perceptual_error_ours}
\end{figure*}

Fig.~\ref{fig:ablation_study-fvs_error-canonical_feature_oflow_ours} shows the reconstruction quality of our method and different modeling choices for a fixed unseen expression and a novel camera viewpoint rendering (a held-out view). Here, the error map (bottom row) represents a pixel-wise mean square error (MSE) of head renderings in RGB color space. Fig.~\ref{fig:ablation_study-dvs-perceptual_error_ours} further compares our approach with the same design choices, for a fixed expression but under dynamic novel viewpoint synthesis. Note that dynamic viewpoints are interpolated from different camera viewpoints.
From these results, we can observe that without conditioning the canonical space on the driving image features 
the reconstruction has blurry artifacts all over the mouth. Without the optical flow based loss, blocky artifacts and/or inconsistent fine-scale details appear in sparsely sampled regions, such as hair, eyelids, and teeth. Note that a canonical space representation is required for proper encoding of facial dynamics; otherwise, artifacts emerge. 
The error heatmap visualization in Fig.~\ref{fig:ablation_study-fvs_error-canonical_feature_oflow_ours} (bottom row) provides a quantitative measurement of the error distribution, showing that our approach with all design choices achieves the best rendering quality. 
Tab.~\ref{tab:quantitative_results-ave_error-ablation} shows the average reconstruction error over the entire test set (200 frames) for different well-established image-based quality metrics.
We adopt similar metrics to that of MVP~\cite{lombardi21_mvp}. 
We measure the Manhattan distance $L1$ in the RGB color space, PSNR, SSIM~\cite{wang2004_image-quality}, and LPIPS~\cite{zhang2018_unreasonable-effectiveness}.
Overall our approach attains the best numerical results. This study confirms that our key modeling choices optimize the rendering quality. 
We also show in Fig.~\ref{fig:ablation_study-details-perceptual_error_ours} that the perceptual loss and error map based sampling improve the rendering results. While we have noticed that these components help in improving rendering quality, we do not emphasize them as a contribution.

\subsection{Comparisons with the State of the Art}\label{sec:comparisons}

\begin{figure*}[htb]
\centering
\includegraphics[width=\textwidth]{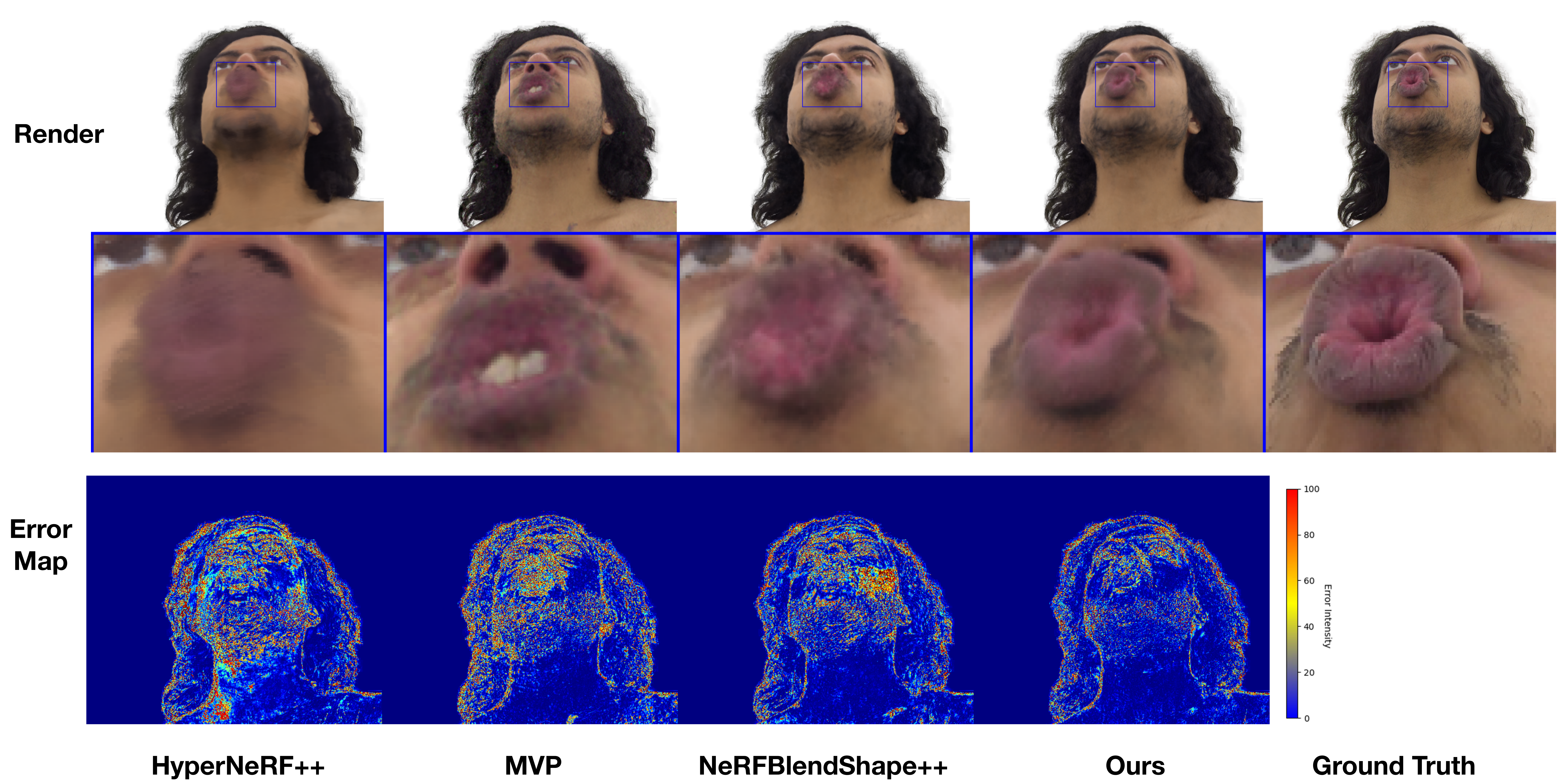}
\caption{Quantitative comparison with the state of the art: \textit{Left to right}: Results of HyperNeRF++~\cite{park2021_hypernerf}, MVP~\cite{lombardi21_mvp}, NeRFBlendShape++~\cite{gao2022_nerfblendshape}, ours and ground truth. The top row shows visual results, while error maps are shown in the bottom row. The error is computed as the per-pixel mean squared error (MSE), encoded in RGB color space. Here, blue denotes $0$ MSE, yellow is $60$ MSE, and reddish colors mean over $100$ MSE. Our method clearly outperforms the state of the art.
}
\label{fig:comparison_stota-fvs_error}
\end{figure*}

\begin{figure*}[htb]
\centering
\includegraphics[width=\textwidth]{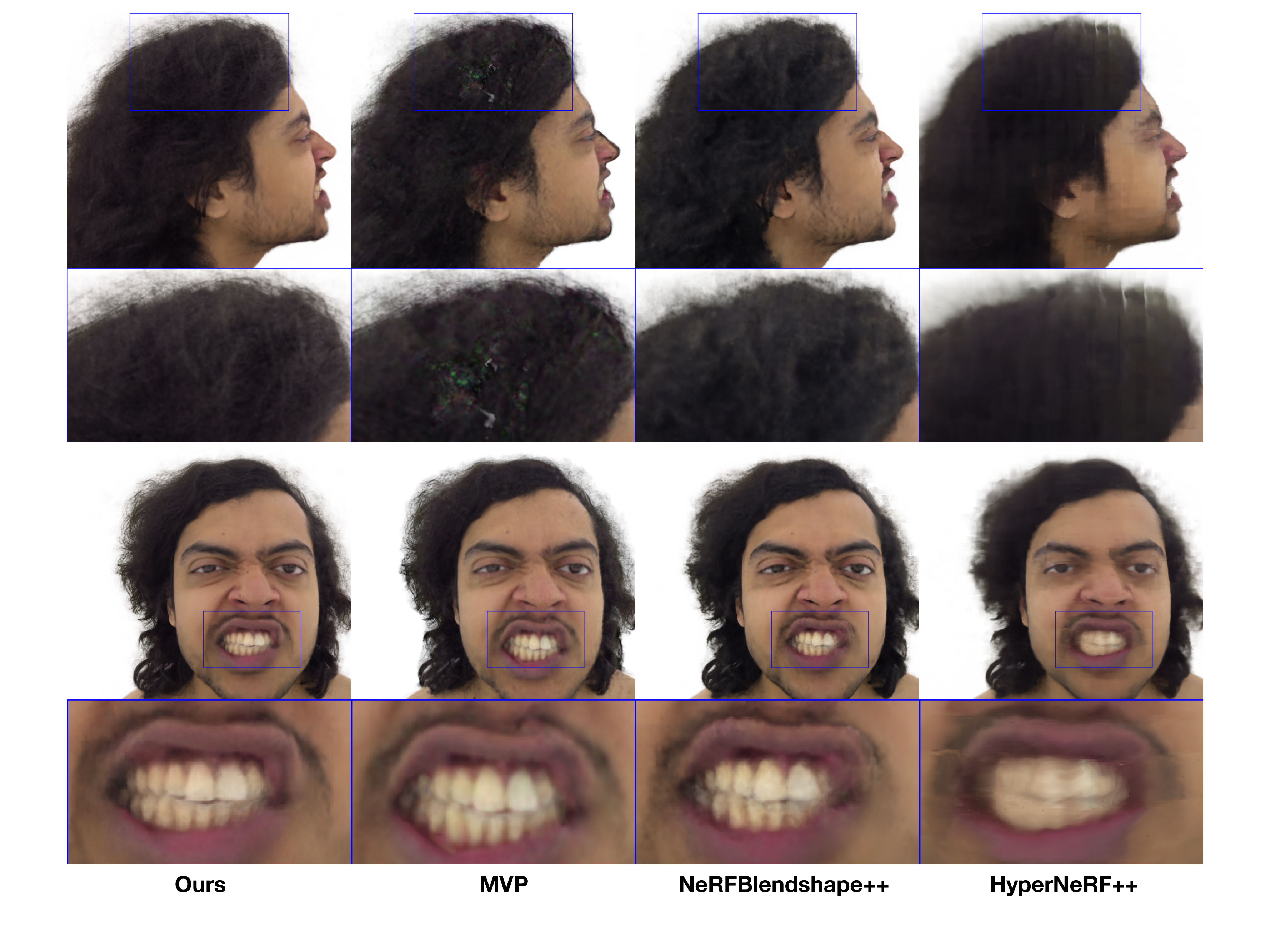}
\caption{Qualitative comparisons with the state of the art in a novel view synthesis setting. \textit{Left to right}: Ours, MVP~\cite{lombardi21_mvp}, NerFBlendshape++~\cite{gao2022_nerfblendshape}, and HyperNeRF++~\cite{park2021_hypernerf}. Unlike other baseline implementations, our approach produces crisper details and more accurate results.}
\label{fig:comparison_stota-fvs}
\end{figure*}
\begin{figure}[]
	\centering
    \includegraphics[width=\linewidth]{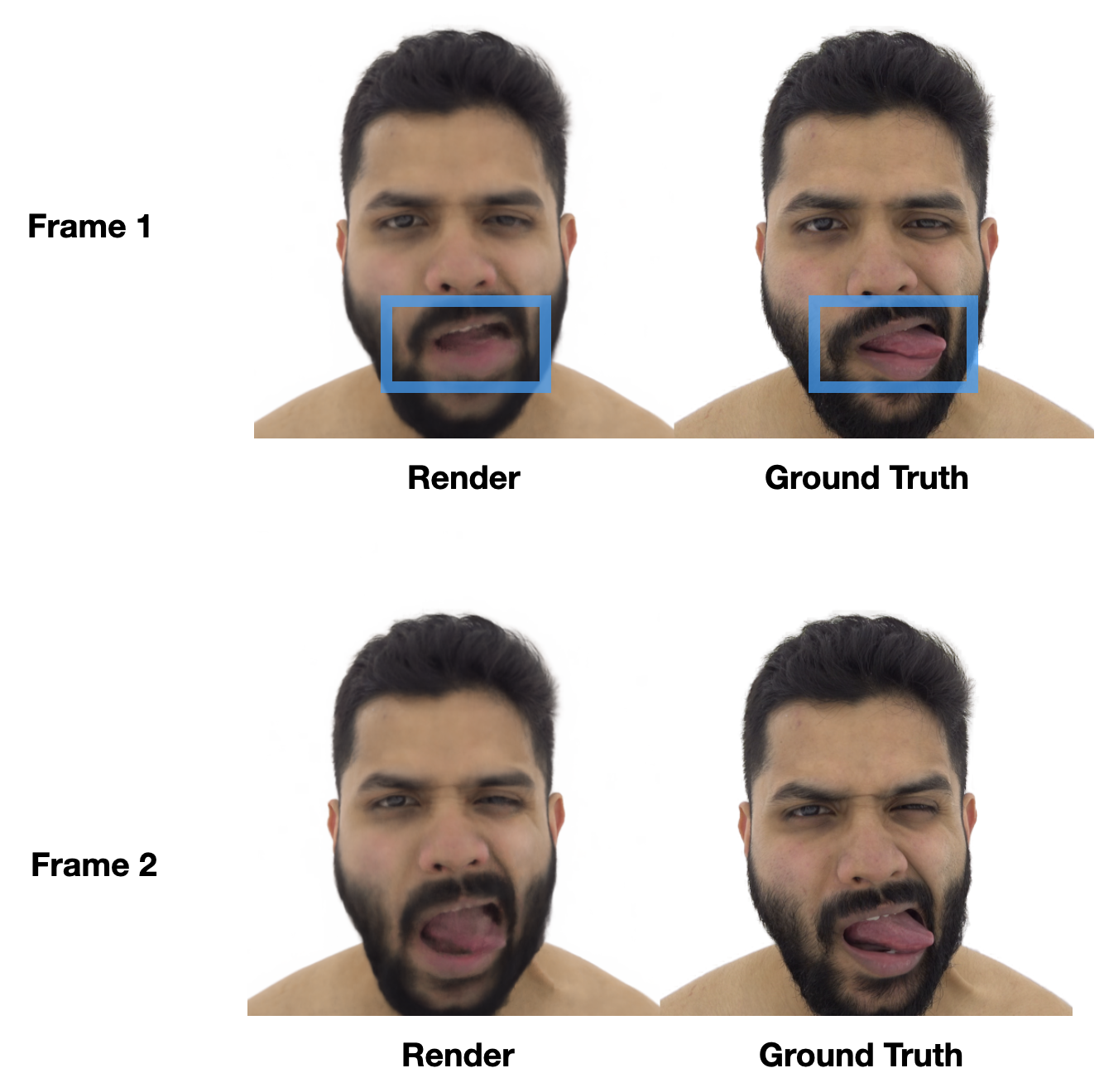}
	\caption{Our method can struggle to faithfully reconstruct the tongue while transitioning from the mouth interior to the outside of the mouth (Frame 1, see blue region). Once the tongue is out, our method captures the tongue with good quality (Frame 2).}
	\label{fig:tonguelimiation}
\end{figure}
In this section, we compare our approach with a recent multi-view state-of-the-art method, called MVP~\cite{lombardi21_mvp}, which produces detailed avatars with high fidelity under a similar setup to ours. We disregard direct comparisons with state-of-the-art sparse multi-view approaches since they tend to lack fine-scale details or are prone to artifacts for novel viewpoint synthesis (see Sec.~\ref{sec:related_work}). In addition, we provide baseline comparisons with an adaptation of a template-free dynamic representation, called HyperNeRF \cite{park2021_hypernerf}, and a multi-level hash table-based approach for expression encoding, called NeRFBlendShape \cite{gao2022_nerfblendshape}. We will call our multi-view and image-driven adaptation of these approaches HyperNeRF++ and NeRFBlendShape++.

To train NeRFBlendShape++, we pass each entry of the expression latent vector to a learnable multi-level hash table. We linearly combine the output of these hash tables and condition the NeRF network on it. To train HyperNeRF++, we feed the neural features passed on by the image encoder to an ambient and deformation network and then as appearance conditioning to the NeRF network.
To run MVP, we use 4k primitives. We employ an in-house FLAME-based tracking to obtain a non-detailed dense reconstruction of the subject's head to guide the initialization of the primitives at each frame.

Fig.~\ref{fig:comparison_stota-fvs_error} shows the reconstruction quality of our method and baseline approaches for a fixed unseen expression and a novel camera viewpoint rendering (a held-out view), while Fig.~\ref{fig:comparison_stota-fvs} compares them in a free-viewpoint synthesis setup. HyperNeRF++ over smooths regions.
Both NeRFBlendShape++ and HyperNeRF++ exhibit artifacts in regions that undergo recurrent topological changes, e.g., the mouth interior, or that have complex structures, e.g., scalp hair. The latter not only produces stronger artifacts in the form of grid patterns but also removes facial details. Overall these methods generalize poorly due to over-parameterized representations.
MVP can sometimes produce wrong facial expressions in extreme cases or even sometimes show unusual block artifacts for the same regions mentioned above. One of the main reasons is that MVP relies on very dense multi-view imagery to supervise volume rendering. However, in a sparser camera setup undersampled areas, especially those undergoing disocclusions, become ambiguous without explicit dense volume deformation constraints. The error heatmap visualization of Fig.~\ref{fig:comparison_stota-fvs_error} (last row), shows that our method reduces reconstruction errors.
Overall our approach produces sharper, more accurate, and more photorealistic rendering results. Please refer to the supplementary video for further comparisons in dynamic viewpoint synthesis.

We perform quantitative evaluations on the 2 held-out views, with 200 frames each. Quantitative comparisons are reported in Tab.~\ref{tab:quantitative_results-ave_error-sota}. Our approach clearly outperforms other baseline approaches, especially when comparing perceptual metrics, such as SSIM and LPIPS. L1 reconstruction error is also significantly reduced.
We remark that our approach attains sharper reconstructions with faster convergence and efficiency, the latter thanks to hash-encoding and empty-space pruning techniques.

\begin{table}[!ht]
\centering
\small{
\begin{tabular}{|c|c|c|c|c|}
\hline
Metrics & HyperNeRF++ & MVP & NeRFBlendshape++ & Ours   \\
\hline
PSNR $\uparrow$  &26.42 & 28.72 &29.66 &\textbf{31.23} \\
\hline
L1 $\downarrow$ &5.61& 3.64& 3.23&  \textbf{2.79}  \\
\hline
SSIM $\uparrow$ &0.8509& 0.8283& 0.8745& \textbf{0.8837}  \\
\hline
LPIPS $\downarrow$ &0.1721& 0.1432&0.1326 & \textbf{0.1130}   \\
\hline
\end{tabular}
}
\caption{Quantitative comparison with state-of-the-art approaches. L1 measures the absolute error of unnormalized RGB images. Our approach outperforms related methods (see bold text).}
\label{tab:quantitative_results-ave_error-sota}
\end{table}

\section{Limitations and Future Work} \label{sec:discussions}

Our method produces highly photorealistic renderings with novel viewpoints and expressions. However, it suffers from a number of limitations. First, we noticed that it can generate artifacts in motions undergoing strong disocclusions (uncovering occlusions). For instance, in the case of the tongue, artifacts could occur around the mouth boundaries as the tongue starts to come out (see Fig.~\ref{fig:tonguelimiation}, Frame 1, blue region). The rendering quality, however, stabilizes with good quality as soon as the tongue becomes fully visible (see Fig.~\ref{fig:tonguelimiation}, Frame 2). Future work could address this limitation e.g. by including occlusion-aware priors.
Second, our solution is currently person-specific. Future work could examine building a model that generalizes to unseen identities. For this, our dataset of 16 identities is a good starting point, though it might require more identities. Here, we could also investigate refining the model using in-the-wild data.
Third, while we have shown real-time renderings at a resolution of $480\times270$, future avenues could enable real-time rendering at higher resolutions e.g. FHD synthesis. Here, we could investigate for instance super-resolution techniques, akin to~\cite{chan2022_efficient-geom-aware,Xiang22GRAM-HD}. 
Finally, we have shown results driven by monocular RGB videos so far. Theoretically, our image encoder could be replaced with other pre-trained encoders of different input modalities, such as audio signals. This would increase the spectrum of applications of our work.

\section{Conclusion} \label{sec:conclusion}

We presented a novel approach for building high-quality digital head avatars using multiresolution hash encoding. Our approach models a full head avatar as a deformation of a canonical space conditioned on the input image.
Our approach utilizes a novel optical flow based loss that enforces correspondences in the learnable canonical space. This encourages artifact-free and temporally smooth results. Our technique is trained in a supervised manner using multi-view RGB data and at inference is driven using monocular input. We have shown results rendered with novel camera viewpoints and expressions. We have also shown different applications including driving the model from novel viewpoints. Our approach also shows the first 2K renderings in literature and can run in real-time at a 480x270 resolution. Overall our approach outperforms all existing methods, both visually and numerically. We will release a novel dataset of 16 identities captured by 24 camera viewpoints and performing a variety of expressions. We hope our work brings human digitization closer to reality so that we all can stay in touch with our friends, family, and loved ones, over a distance.

\bibliographystyle{ACM-Reference-Format}
\bibliography{references}


\begin{thebibliography}{69}

%%% the Web address.

\ifx \showCODEN    \undefined \def \showCODEN     #1{\unskip}     \fi
\ifx \showDOI      \undefined \def \showDOI       #1{#1}\fi
\ifx \showISBNx    \undefined \def \showISBNx     #1{\unskip}     \fi
\ifx \showISBNxiii \undefined \def \showISBNxiii  #1{\unskip}     \fi
\ifx \showISSN     \undefined \def \showISSN      #1{\unskip}     \fi
\ifx \showLCCN     \undefined \def \showLCCN      #1{\unskip}     \fi
\ifx \shownote     \undefined \def \shownote      #1{#1}          \fi
\ifx \showarticletitle \undefined \def \showarticletitle #1{#1}   \fi
\ifx \showURL      \undefined \def \showURL       {\relax}        \fi
% invisible to TeX
\providecommand\bibfield[2]{#2}
\providecommand\bibinfo[2]{#2}
\providecommand\natexlab[1]{#1}
\providecommand\showeprint[2][]{arXiv:#2}

\bibitem[\protect\citeauthoryear{Abdal, Lee, Zhu, Chai, Siarohin, Wonka, and
  Tulyakov}{Abdal et~al\mbox{.}}{2023}]%
        {abdal2023_3davatargan}
\bibfield{author}{\bibinfo{person}{Rameen Abdal}, \bibinfo{person}{Hsin{-}Ying
  Lee}, \bibinfo{person}{Peihao Zhu}, \bibinfo{person}{Menglei Chai},
  \bibinfo{person}{Aliaksandr Siarohin}, \bibinfo{person}{Peter Wonka}, {and}
  \bibinfo{person}{Sergey Tulyakov}.} \bibinfo{year}{2023}\natexlab{}.
\newblock \showarticletitle{3DAvatarGAN: Bridging Domains for Personalized
  Editable Avatars}.
\newblock \bibinfo{journal}{\emph{CoRR}}  \bibinfo{volume}{abs/2301.02700}
  (\bibinfo{year}{2023}).
\newblock


\bibitem[\protect\citeauthoryear{Athar, Xu, Sunkavalli, Shechtman, and
  Shu}{Athar et~al\mbox{.}}{2022}]%
        {athar2022rignerf}
\bibfield{author}{\bibinfo{person}{ShahRukh Athar}, \bibinfo{person}{Zexiang
  Xu}, \bibinfo{person}{Kalyan Sunkavalli}, \bibinfo{person}{Eli Shechtman},
  {and} \bibinfo{person}{Zhixin Shu}.} \bibinfo{year}{2022}\natexlab{}.
\newblock \showarticletitle{RigNeRF: Fully Controllable Neural 3D Portraits}.
  In \bibinfo{booktitle}{\emph{IEEE Conf. Comput. Vis. Pattern Recog.}}
  \bibinfo{publisher}{{IEEE}}, \bibinfo{pages}{20332--20341}.
\newblock


\bibitem[\protect\citeauthoryear{Bai, Fan, Wang, Zhang, Sun, Yuan, and
  Shan}{Bai et~al\mbox{.}}{2022}]%
        {bai2022_high-fidelity}
\bibfield{author}{\bibinfo{person}{Yunpeng Bai}, \bibinfo{person}{Yanbo Fan},
  \bibinfo{person}{Xuan Wang}, \bibinfo{person}{Yong Zhang},
  \bibinfo{person}{Jingxiang Sun}, \bibinfo{person}{Chun Yuan}, {and}
  \bibinfo{person}{Ying Shan}.} \bibinfo{year}{2022}\natexlab{}.
\newblock \showarticletitle{High-fidelity Facial Avatar Reconstruction from
  Monocular Video with Generative Priors}.
\newblock \bibinfo{journal}{\emph{CoRR}}  \bibinfo{volume}{abs/2211.15064}
  (\bibinfo{year}{2022}).
\newblock


\bibitem[\protect\citeauthoryear{B.R., Tewari, Seidel, Elgharib, and
  Theobalt}{B.R. et~al\mbox{.}}{2021}]%
        {mallikarjun2021_learning-complete}
\bibfield{author}{\bibinfo{person}{Mallikarjun B.R.}, \bibinfo{person}{Ayush
  Tewari}, \bibinfo{person}{Hans{-}Peter Seidel}, \bibinfo{person}{Mohamed
  Elgharib}, {and} \bibinfo{person}{Christian Theobalt}.}
  \bibinfo{year}{2021}\natexlab{}.
\newblock \showarticletitle{Learning Complete 3D Morphable Face Models From
  Images and Videos}. In \bibinfo{booktitle}{\emph{IEEE Conf. Comput. Vis.
  Pattern Recog.}} \bibinfo{publisher}{Computer Vision Foundation / {IEEE}
  Computer Society}, \bibinfo{pages}{3361--3371}.
\newblock


\bibitem[\protect\citeauthoryear{Cao, Weng, Zhou, Tong, and Zhou}{Cao
  et~al\mbox{.}}{2014}]%
        {cao2014_facewarehouse}
\bibfield{author}{\bibinfo{person}{Chen Cao}, \bibinfo{person}{Yanlin Weng},
  \bibinfo{person}{Shun Zhou}, \bibinfo{person}{Yiying Tong}, {and}
  \bibinfo{person}{Kun Zhou}.} \bibinfo{year}{2014}\natexlab{}.
\newblock \showarticletitle{FaceWarehouse: {A} 3D Facial Expression Database
  for Visual Computing}.
\newblock \bibinfo{journal}{\emph{IEEE Trans. Vis. Comput. Graph.}}
  \bibinfo{volume}{20}, \bibinfo{number}{3} (\bibinfo{year}{2014}),
  \bibinfo{pages}{413--425}.
\newblock


\bibitem[\protect\citeauthoryear{Cao, Wu, Weng, Shao, and Zhou}{Cao
  et~al\mbox{.}}{2016}]%
        {cao2016_realtime}
\bibfield{author}{\bibinfo{person}{Chen Cao}, \bibinfo{person}{Hongzhi Wu},
  \bibinfo{person}{Yanlin Weng}, \bibinfo{person}{Tianjia Shao}, {and}
  \bibinfo{person}{Kun Zhou}.} \bibinfo{year}{2016}\natexlab{}.
\newblock \showarticletitle{Real-time facial animation with image-based dynamic
  avatars}.
\newblock \bibinfo{journal}{\emph{ACM Trans. Graph.}} \bibinfo{volume}{35},
  \bibinfo{number}{4} (\bibinfo{year}{2016}), \bibinfo{pages}{126:1--126:12}.
\newblock


\bibitem[\protect\citeauthoryear{Chan, Lin, Chan, Nagano, Pan, Mello, Gallo,
  Guibas, Tremblay, Khamis, Karras, and Wetzstein}{Chan et~al\mbox{.}}{2022}]%
        {chan2022_efficient-geom-aware}
\bibfield{author}{\bibinfo{person}{Eric~R. Chan}, \bibinfo{person}{Connor~Z.
  Lin}, \bibinfo{person}{Matthew~A. Chan}, \bibinfo{person}{Koki Nagano},
  \bibinfo{person}{Boxiao Pan}, \bibinfo{person}{Shalini~De Mello},
  \bibinfo{person}{Orazio Gallo}, \bibinfo{person}{Leonidas~J. Guibas},
  \bibinfo{person}{Jonathan Tremblay}, \bibinfo{person}{Sameh Khamis},
  \bibinfo{person}{Tero Karras}, {and} \bibinfo{person}{Gordon Wetzstein}.}
  \bibinfo{year}{2022}\natexlab{}.
\newblock \showarticletitle{Efficient Geometry-aware 3D Generative Adversarial
  Networks}. In \bibinfo{booktitle}{\emph{IEEE Conf. Comput. Vis. Pattern
  Recog.}} \bibinfo{publisher}{{IEEE}}, \bibinfo{pages}{16102--16112}.
\newblock


\bibitem[\protect\citeauthoryear{Chan, Monteiro, Kellnhofer, Wu, and
  Wetzstein}{Chan et~al\mbox{.}}{2021}]%
        {chan2021_pigan}
\bibfield{author}{\bibinfo{person}{Eric~R. Chan}, \bibinfo{person}{Marco
  Monteiro}, \bibinfo{person}{Petr Kellnhofer}, \bibinfo{person}{Jiajun Wu},
  {and} \bibinfo{person}{Gordon Wetzstein}.} \bibinfo{year}{2021}\natexlab{}.
\newblock \showarticletitle{Pi-GAN: Periodic Implicit Generative Adversarial
  Networks for 3D-Aware Image Synthesis}. In \bibinfo{booktitle}{\emph{IEEE
  Conf. Comput. Vis. Pattern Recog.}} \bibinfo{publisher}{Computer Vision
  Foundation / {IEEE} Computer Society}, \bibinfo{pages}{5799--5809}.
\newblock


\bibitem[\protect\citeauthoryear{Chandran, Winberg, Zoss, Riviere, Gross,
  Gotardo, and Bradley}{Chandran et~al\mbox{.}}{2021}]%
        {chandran2021_rendering-style}
\bibfield{author}{\bibinfo{person}{Prashanth Chandran},
  \bibinfo{person}{Sebastian Winberg}, \bibinfo{person}{Gaspard Zoss},
  \bibinfo{person}{J{\'{e}}r{\'{e}}my Riviere}, \bibinfo{person}{Markus~H.
  Gross}, \bibinfo{person}{Paulo F.~U. Gotardo}, {and} \bibinfo{person}{Derek
  Bradley}.} \bibinfo{year}{2021}\natexlab{}.
\newblock \showarticletitle{Rendering with style: combining traditional and
  neural approaches for high-quality face rendering}.
\newblock \bibinfo{journal}{\emph{ACM Trans. Graph.}} \bibinfo{volume}{40},
  \bibinfo{number}{6} (\bibinfo{year}{2021}), \bibinfo{pages}{223:1--223:14}.
\newblock


\bibitem[\protect\citeauthoryear{Deng, Yang, Xiang, and Tong}{Deng
  et~al\mbox{.}}{2022}]%
        {deng2022_gram}
\bibfield{author}{\bibinfo{person}{Yu Deng}, \bibinfo{person}{Jiaolong Yang},
  \bibinfo{person}{Jianfeng Xiang}, {and} \bibinfo{person}{Xin Tong}.}
  \bibinfo{year}{2022}\natexlab{}.
\newblock \showarticletitle{{GRAM:} Generative Radiance Manifolds for 3D-Aware
  Image Generation}. In \bibinfo{booktitle}{\emph{IEEE Conf. Comput. Vis.
  Pattern Recog.}} \bibinfo{publisher}{{IEEE}}, \bibinfo{pages}{10663--10673}.
\newblock


\bibitem[\protect\citeauthoryear{Egger, Smith, Tewari, Wuhrer, Zollh{\"{o}}fer,
  Beeler, Bernard, Bolkart, Kortylewski, Romdhani, Theobalt, Blanz, and
  Vetter}{Egger et~al\mbox{.}}{2020}]%
        {egger2020_3dmfm}
\bibfield{author}{\bibinfo{person}{Bernhard Egger}, \bibinfo{person}{William
  A.~P. Smith}, \bibinfo{person}{Ayush Tewari}, \bibinfo{person}{Stefanie
  Wuhrer}, \bibinfo{person}{Michael Zollh{\"{o}}fer}, \bibinfo{person}{Thabo
  Beeler}, \bibinfo{person}{Florian Bernard}, \bibinfo{person}{Timo Bolkart},
  \bibinfo{person}{Adam Kortylewski}, \bibinfo{person}{Sami Romdhani},
  \bibinfo{person}{Christian Theobalt}, \bibinfo{person}{Volker Blanz}, {and}
  \bibinfo{person}{Thomas Vetter}.} \bibinfo{year}{2020}\natexlab{}.
\newblock \showarticletitle{3D Morphable Face Models - Past, Present, and
  Future}.
\newblock \bibinfo{journal}{\emph{ACM Trans. Graph.}} \bibinfo{volume}{39},
  \bibinfo{number}{5} (\bibinfo{year}{2020}), \bibinfo{pages}{157:1--157:38}.
\newblock


\bibitem[\protect\citeauthoryear{Fridovich{-}Keil, Yu, Tancik, Chen, Recht, and
  Kanazawa}{Fridovich{-}Keil et~al\mbox{.}}{2022}]%
        {fridovich2022_plenoxels}
\bibfield{author}{\bibinfo{person}{Sara Fridovich{-}Keil},
  \bibinfo{person}{Alex Yu}, \bibinfo{person}{Matthew Tancik},
  \bibinfo{person}{Qinhong Chen}, \bibinfo{person}{Benjamin Recht}, {and}
  \bibinfo{person}{Angjoo Kanazawa}.} \bibinfo{year}{2022}\natexlab{}.
\newblock \showarticletitle{Plenoxels: Radiance Fields without Neural
  Networks}. In \bibinfo{booktitle}{\emph{IEEE Conf. Comput. Vis. Pattern
  Recog.}} \bibinfo{publisher}{{IEEE}}, \bibinfo{pages}{5491--5500}.
\newblock


\bibitem[\protect\citeauthoryear{Gafni, Thies, Zollh{\"o}fer, and
  Nie{\ss}ner}{Gafni et~al\mbox{.}}{2021}]%
        {gafni2021_dynamic-nerf}
\bibfield{author}{\bibinfo{person}{Guy Gafni}, \bibinfo{person}{Justus Thies},
  \bibinfo{person}{Michael Zollh{\"o}fer}, {and} \bibinfo{person}{Matthias
  Nie{\ss}ner}.} \bibinfo{year}{2021}\natexlab{}.
\newblock \showarticletitle{Dynamic Neural Radiance Fields for Monocular 4D
  Facial Avatar Reconstruction}. In \bibinfo{booktitle}{\emph{IEEE Conf.
  Comput. Vis. Pattern Recog.}} \bibinfo{publisher}{{IEEE}},
  \bibinfo{pages}{8649--8658}.
\newblock


\bibitem[\protect\citeauthoryear{Gao, Zhong, Xiang, Hong, Guo, and Zhang}{Gao
  et~al\mbox{.}}{2022}]%
        {gao2022_nerfblendshape}
\bibfield{author}{\bibinfo{person}{Xuan Gao}, \bibinfo{person}{Chenglai Zhong},
  \bibinfo{person}{Jun Xiang}, \bibinfo{person}{Yang Hong},
  \bibinfo{person}{Yudong Guo}, {and} \bibinfo{person}{Juyong Zhang}.}
  \bibinfo{year}{2022}\natexlab{}.
\newblock \showarticletitle{Reconstructing Personalized Semantic Facial NeRF
  Models from Monocular Video}.
\newblock \bibinfo{journal}{\emph{ACM Trans. Graph.}} \bibinfo{volume}{41},
  \bibinfo{number}{6} (\bibinfo{year}{2022}), \bibinfo{pages}{200:1--200:12}.
\newblock


\bibitem[\protect\citeauthoryear{Gecer, Ploumpis, Kotsia, and Zafeiriou}{Gecer
  et~al\mbox{.}}{2019}]%
        {grecer2019_ganfit}
\bibfield{author}{\bibinfo{person}{Baris Gecer}, \bibinfo{person}{Stylianos
  Ploumpis}, \bibinfo{person}{Irene Kotsia}, {and} \bibinfo{person}{Stefanos
  Zafeiriou}.} \bibinfo{year}{2019}\natexlab{}.
\newblock \showarticletitle{{GANFIT:} Generative Adversarial Network Fitting
  for High Fidelity 3D Face Reconstruction}. In \bibinfo{booktitle}{\emph{IEEE
  Conf. Comput. Vis. Pattern Recog.}} \bibinfo{publisher}{Computer Vision
  Foundation / {IEEE} Computer Society}, \bibinfo{pages}{1155--1164}.
\newblock


\bibitem[\protect\citeauthoryear{Gerig, Morel{-}Forster, Blumer, Egger,
  L{\"{u}}thi, Sch{\"{o}}nborn, and Vetter}{Gerig et~al\mbox{.}}{2018}]%
        {gerig2018_morphable_models}
\bibfield{author}{\bibinfo{person}{Thomas Gerig}, \bibinfo{person}{Andreas
  Morel{-}Forster}, \bibinfo{person}{Clemens Blumer}, \bibinfo{person}{Bernhard
  Egger}, \bibinfo{person}{Marcel L{\"{u}}thi}, \bibinfo{person}{Sandro
  Sch{\"{o}}nborn}, {and} \bibinfo{person}{Thomas Vetter}.}
  \bibinfo{year}{2018}\natexlab{}.
\newblock \showarticletitle{Morphable Face Models - An Open Framework}. In
  \bibinfo{booktitle}{\emph{Conference on Automatic Face {\&} Gesture
  Recognition}}. \bibinfo{publisher}{{IEEE} Computer Society},
  \bibinfo{pages}{75--82}.
\newblock


\bibitem[\protect\citeauthoryear{Grassal, Prinzler, Leistner, Rother,
  Nie{\ss}ner, and Thies}{Grassal et~al\mbox{.}}{2022}]%
        {grassal2022_neural-head}
\bibfield{author}{\bibinfo{person}{Philip{-}William Grassal},
  \bibinfo{person}{Malte Prinzler}, \bibinfo{person}{Titus Leistner},
  \bibinfo{person}{Carsten Rother}, \bibinfo{person}{Matthias Nie{\ss}ner},
  {and} \bibinfo{person}{Justus Thies}.} \bibinfo{year}{2022}\natexlab{}.
\newblock \showarticletitle{Neural Head Avatars from Monocular {RGB} Videos}.
  In \bibinfo{booktitle}{\emph{IEEE Conf. Comput. Vis. Pattern Recog.}}
  \bibinfo{publisher}{{IEEE}}, \bibinfo{pages}{18632--18643}.
\newblock


\bibitem[\protect\citeauthoryear{Gu, Liu, Wang, and Theobalt}{Gu
  et~al\mbox{.}}{2022}]%
        {gu2022_stylenerf}
\bibfield{author}{\bibinfo{person}{Jiatao Gu}, \bibinfo{person}{Lingjie Liu},
  \bibinfo{person}{Peng Wang}, {and} \bibinfo{person}{Christian Theobalt}.}
  \bibinfo{year}{2022}\natexlab{}.
\newblock \showarticletitle{StyleNeRF: {A} Style-based 3D Aware Generator for
  High-resolution Image Synthesis}. In \bibinfo{booktitle}{\emph{Int. Conf.
  Learn. Represent.}} \bibinfo{publisher}{OpenReview.net}.
\newblock


\bibitem[\protect\citeauthoryear{Hong, Peng, Xiao, Liu, and Zhang}{Hong
  et~al\mbox{.}}{2022}]%
        {hong2021_headnerf}
\bibfield{author}{\bibinfo{person}{Yang Hong}, \bibinfo{person}{Bo Peng},
  \bibinfo{person}{Haiyao Xiao}, \bibinfo{person}{Ligang Liu}, {and}
  \bibinfo{person}{Juyong Zhang}.} \bibinfo{year}{2022}\natexlab{}.
\newblock \showarticletitle{HeadNeRF: {A} Realtime NeRF-based Parametric Head
  Model}. In \bibinfo{booktitle}{\emph{IEEE Conf. Comput. Vis. Pattern Recog.}}
  \bibinfo{publisher}{{IEEE}}, \bibinfo{pages}{20342--20352}.
\newblock


\bibitem[\protect\citeauthoryear{Ichim, Bouaziz, and Pauly}{Ichim
  et~al\mbox{.}}{2015}]%
        {ichim2015_dynamic_3davatar}
\bibfield{author}{\bibinfo{person}{Alexandru~Eugen Ichim},
  \bibinfo{person}{Sofien Bouaziz}, {and} \bibinfo{person}{Mark Pauly}.}
  \bibinfo{year}{2015}\natexlab{}.
\newblock \showarticletitle{Dynamic 3D avatar creation from hand-held video
  input}.
\newblock \bibinfo{journal}{\emph{ACM Trans. Graph.}} \bibinfo{volume}{34},
  \bibinfo{number}{4} (\bibinfo{year}{2015}), \bibinfo{pages}{45:1--45:14}.
\newblock


\bibitem[\protect\citeauthoryear{Kasten, Ofri, Wang, and Dekel}{Kasten
  et~al\mbox{.}}{2021}]%
        {kasten2021layered}
\bibfield{author}{\bibinfo{person}{Yoni Kasten}, \bibinfo{person}{Dolev Ofri},
  \bibinfo{person}{Oliver Wang}, {and} \bibinfo{person}{Tali Dekel}.}
  \bibinfo{year}{2021}\natexlab{}.
\newblock \showarticletitle{Layered neural atlases for consistent video
  editing}.
\newblock \bibinfo{journal}{\emph{ACM Transactions on Graphics (TOG)}}
  \bibinfo{volume}{40}, \bibinfo{number}{6} (\bibinfo{year}{2021}),
  \bibinfo{pages}{1--12}.
\newblock


\bibitem[\protect\citeauthoryear{Kim, Garrido, Tewari, Xu, Thies, Nie{\ss}ner,
  P{\'{e}}rez, Richardt, Zollh{\"{o}}fer, and Theobalt}{Kim
  et~al\mbox{.}}{2018}]%
        {kim2018_dvp}
\bibfield{author}{\bibinfo{person}{Hyeongwoo Kim}, \bibinfo{person}{Pablo
  Garrido}, \bibinfo{person}{Ayush Tewari}, \bibinfo{person}{Weipeng Xu},
  \bibinfo{person}{Justus Thies}, \bibinfo{person}{Matthias Nie{\ss}ner},
  \bibinfo{person}{Patrick P{\'{e}}rez}, \bibinfo{person}{Christian Richardt},
  \bibinfo{person}{Michael Zollh{\"{o}}fer}, {and} \bibinfo{person}{Christian
  Theobalt}.} \bibinfo{year}{2018}\natexlab{}.
\newblock \showarticletitle{Deep video portraits}.
\newblock \bibinfo{journal}{\emph{ACM Trans. Graph.}} \bibinfo{volume}{37},
  \bibinfo{number}{4} (\bibinfo{year}{2018}), \bibinfo{pages}{163}.
\newblock


\bibitem[\protect\citeauthoryear{Lattas, Moschoglou, Ploumpis, Gecer, Ghosh,
  and Zafeiriou}{Lattas et~al\mbox{.}}{2022}]%
        {lattas2022_avatarme++}
\bibfield{author}{\bibinfo{person}{Alexandros Lattas},
  \bibinfo{person}{Stylianos Moschoglou}, \bibinfo{person}{Stylianos Ploumpis},
  \bibinfo{person}{Baris Gecer}, \bibinfo{person}{Abhijeet Ghosh}, {and}
  \bibinfo{person}{Stefanos Zafeiriou}.} \bibinfo{year}{2022}\natexlab{}.
\newblock \showarticletitle{AvatarMe\({}^{\mbox{++}}\): Facial Shape and {BRDF}
  Inference With Photorealistic Rendering-Aware GANs}.
\newblock \bibinfo{journal}{\emph{IEEE Trans. Pattern Anal. Mach. Intell.}}
  \bibinfo{volume}{44}, \bibinfo{number}{12} (\bibinfo{year}{2022}),
  \bibinfo{pages}{9269--9284}.
\newblock


\bibitem[\protect\citeauthoryear{Li, Tanke, Vo, Zollh{\"{o}}fer, Gall,
  Kanazawa, and Lassner}{Li et~al\mbox{.}}{2022b}]%
        {li2022_tava}
\bibfield{author}{\bibinfo{person}{Ruilong Li}, \bibinfo{person}{Julian Tanke},
  \bibinfo{person}{Minh Vo}, \bibinfo{person}{Michael Zollh{\"{o}}fer},
  \bibinfo{person}{J{\"{u}}rgen Gall}, \bibinfo{person}{Angjoo Kanazawa}, {and}
  \bibinfo{person}{Christoph Lassner}.} \bibinfo{year}{2022}\natexlab{b}.
\newblock \showarticletitle{{TAVA:} Template-free Animatable Volumetric
  Actors}. In \bibinfo{booktitle}{\emph{Eur. Conf. Comput. Vis.}}
  \emph{(\bibinfo{series}{Lecture Notes in Computer Science},
  Vol.~\bibinfo{volume}{13692})}. \bibinfo{publisher}{Springer},
  \bibinfo{pages}{419--436}.
\newblock


\bibitem[\protect\citeauthoryear{Li, Bolkart, Black, Li, and Romero}{Li
  et~al\mbox{.}}{2017}]%
        {li2017_flame}
\bibfield{author}{\bibinfo{person}{Tianye Li}, \bibinfo{person}{Timo Bolkart},
  \bibinfo{person}{Michael~J. Black}, \bibinfo{person}{Hao Li}, {and}
  \bibinfo{person}{Javier Romero}.} \bibinfo{year}{2017}\natexlab{}.
\newblock \showarticletitle{Learning a model of facial shape and expression
  from 4D scans}.
\newblock \bibinfo{journal}{\emph{ACM Trans. Graph.}} \bibinfo{volume}{36},
  \bibinfo{number}{6} (\bibinfo{year}{2017}), \bibinfo{pages}{194:1--194:17}.
\newblock


\bibitem[\protect\citeauthoryear{Li, Slavcheva, Zollh{\"{o}}fer, Green,
  Lassner, Kim, Schmidt, Lovegrove, Goesele, Newcombe, and Lv}{Li
  et~al\mbox{.}}{2022a}]%
        {li2022_neural-3d-video}
\bibfield{author}{\bibinfo{person}{Tianye Li}, \bibinfo{person}{Mira
  Slavcheva}, \bibinfo{person}{Michael Zollh{\"{o}}fer}, \bibinfo{person}{Simon
  Green}, \bibinfo{person}{Christoph Lassner}, \bibinfo{person}{Changil Kim},
  \bibinfo{person}{Tanner Schmidt}, \bibinfo{person}{Steven Lovegrove},
  \bibinfo{person}{Michael Goesele}, \bibinfo{person}{Richard~A. Newcombe},
  {and} \bibinfo{person}{Zhaoyang Lv}.} \bibinfo{year}{2022}\natexlab{a}.
\newblock \showarticletitle{Neural 3D Video Synthesis from Multi-view Video}.
  In \bibinfo{booktitle}{\emph{IEEE Conf. Comput. Vis. Pattern Recog.}}
  \bibinfo{publisher}{{IEEE}}, \bibinfo{pages}{5511--5521}.
\newblock


\bibitem[\protect\citeauthoryear{Li, Niklaus, Snavely, and Wang}{Li
  et~al\mbox{.}}{2021}]%
        {li2021_neural-scene}
\bibfield{author}{\bibinfo{person}{Zhengqi Li}, \bibinfo{person}{Simon
  Niklaus}, \bibinfo{person}{Noah Snavely}, {and} \bibinfo{person}{Oliver
  Wang}.} \bibinfo{year}{2021}\natexlab{}.
\newblock \showarticletitle{Neural Scene Flow Fields for Space-Time View
  Synthesis of Dynamic Scenes}. In \bibinfo{booktitle}{\emph{IEEE Conf. Comput.
  Vis. Pattern Recog.}} \bibinfo{publisher}{Computer Vision Foundation / {IEEE}
  Computer Society}, \bibinfo{pages}{6498--6508}.
\newblock


\bibitem[\protect\citeauthoryear{Lin, Yuan, Shao, and Zhou}{Lin
  et~al\mbox{.}}{2020}]%
        {lin2020_towards-hf}
\bibfield{author}{\bibinfo{person}{Jiangke Lin}, \bibinfo{person}{Yi Yuan},
  \bibinfo{person}{Tianjia Shao}, {and} \bibinfo{person}{Kun Zhou}.}
  \bibinfo{year}{2020}\natexlab{}.
\newblock \showarticletitle{Towards High-Fidelity 3D Face Reconstruction From
  In-the-Wild Images Using Graph Convolutional Networks}. In
  \bibinfo{booktitle}{\emph{IEEE Conf. Comput. Vis. Pattern Recog.}}
  \bibinfo{publisher}{Computer Vision Foundation / {IEEE} Computer Society},
  \bibinfo{pages}{5890--5899}.
\newblock


\bibitem[\protect\citeauthoryear{Lin, Ryabtsev, Sengupta, Curless, Seitz, and
  Kemelmacher{-}Shlizerman}{Lin et~al\mbox{.}}{2021}]%
        {lin2021_back_mattv2}
\bibfield{author}{\bibinfo{person}{Shanchuan Lin}, \bibinfo{person}{Andrey
  Ryabtsev}, \bibinfo{person}{Soumyadip Sengupta}, \bibinfo{person}{Brian~L.
  Curless}, \bibinfo{person}{Steven~M. Seitz}, {and} \bibinfo{person}{Ira
  Kemelmacher{-}Shlizerman}.} \bibinfo{year}{2021}\natexlab{}.
\newblock \showarticletitle{Real-Time High-Resolution Background Matting}. In
  \bibinfo{booktitle}{\emph{IEEE Conf. Comput. Vis. Pattern Recog.}}
  \bibinfo{publisher}{Computer Vision Foundation / {IEEE} Computer Society},
  \bibinfo{pages}{8762--8771}.
\newblock


\bibitem[\protect\citeauthoryear{Lombardi, Saragih, Simon, and Sheikh}{Lombardi
  et~al\mbox{.}}{2018}]%
        {lombardi2018_deep-appear}
\bibfield{author}{\bibinfo{person}{Stephen Lombardi}, \bibinfo{person}{Jason~M.
  Saragih}, \bibinfo{person}{Tomas Simon}, {and} \bibinfo{person}{Yaser
  Sheikh}.} \bibinfo{year}{2018}\natexlab{}.
\newblock \showarticletitle{Deep appearance models for face rendering}.
\newblock \bibinfo{journal}{\emph{ACM Trans. Graph.}} \bibinfo{volume}{37},
  \bibinfo{number}{4} (\bibinfo{year}{2018}), \bibinfo{pages}{68}.
\newblock


\bibitem[\protect\citeauthoryear{Lombardi, Simon, Saragih, Schwartz, Lehrmann,
  and Sheikh}{Lombardi et~al\mbox{.}}{2019}]%
        {lombardi2019_neural-volumes}
\bibfield{author}{\bibinfo{person}{Stephen Lombardi}, \bibinfo{person}{Tomas
  Simon}, \bibinfo{person}{Jason Saragih}, \bibinfo{person}{Gabriel Schwartz},
  \bibinfo{person}{Andreas Lehrmann}, {and} \bibinfo{person}{Yaser Sheikh}.}
  \bibinfo{year}{2019}\natexlab{}.
\newblock \showarticletitle{Neural Volumes: Learning Dynamic Renderable Volumes
  from Images}.
\newblock \bibinfo{journal}{\emph{ACM Trans. Graph.}} \bibinfo{volume}{38},
  \bibinfo{number}{4} (\bibinfo{year}{2019}), \bibinfo{pages}{65:1--65:14}.
\newblock


\bibitem[\protect\citeauthoryear{Lombardi, Simon, Schwartz, Zollh{\"{o}}fer,
  Sheikh, and Saragih}{Lombardi et~al\mbox{.}}{2021}]%
        {lombardi21_mvp}
\bibfield{author}{\bibinfo{person}{Stephen Lombardi}, \bibinfo{person}{Tomas
  Simon}, \bibinfo{person}{Gabriel Schwartz}, \bibinfo{person}{Michael
  Zollh{\"{o}}fer}, \bibinfo{person}{Yaser Sheikh}, {and}
  \bibinfo{person}{Jason~M. Saragih}.} \bibinfo{year}{2021}\natexlab{}.
\newblock \showarticletitle{Mixture of volumetric primitives for efficient
  neural rendering}.
\newblock \bibinfo{journal}{\emph{ACM Trans. Graph.}} \bibinfo{volume}{40},
  \bibinfo{number}{4} (\bibinfo{year}{2021}), \bibinfo{pages}{59:1--59:13}.
\newblock


\bibitem[\protect\citeauthoryear{Ma, Simon, Saragih, Wang, Li, la~Torre, and
  Sheikh}{Ma et~al\mbox{.}}{2021}]%
        {ma2021_pixel-codec}
\bibfield{author}{\bibinfo{person}{Shugao Ma}, \bibinfo{person}{Tomas Simon},
  \bibinfo{person}{Jason~M. Saragih}, \bibinfo{person}{Dawei Wang},
  \bibinfo{person}{Yuecheng Li}, \bibinfo{person}{Fernando~De la Torre}, {and}
  \bibinfo{person}{Yaser Sheikh}.} \bibinfo{year}{2021}\natexlab{}.
\newblock \showarticletitle{Pixel Codec Avatars}. In
  \bibinfo{booktitle}{\emph{IEEE Conf. Comput. Vis. Pattern Recog.}}
  \bibinfo{publisher}{Computer Vision Foundation / {IEEE} Computer Society},
  \bibinfo{pages}{64--73}.
\newblock


\bibitem[\protect\citeauthoryear{Meshry, Suri, Davis, and Shrivastava}{Meshry
  et~al\mbox{.}}{2021}]%
        {meshry2021_learned-spatial}
\bibfield{author}{\bibinfo{person}{Moustafa Meshry}, \bibinfo{person}{Saksham
  Suri}, \bibinfo{person}{Larry~S. Davis}, {and} \bibinfo{person}{Abhinav
  Shrivastava}.} \bibinfo{year}{2021}\natexlab{}.
\newblock \showarticletitle{Learned Spatial Representations for Few-shot
  Talking-Head Synthesis}. In \bibinfo{booktitle}{\emph{Int. Conf. Comput.
  Vis.}} \bibinfo{publisher}{{IEEE}}, \bibinfo{pages}{13809--13818}.
\newblock


\bibitem[\protect\citeauthoryear{Metashape}{Metashape}{2020}]%
        {metashape}
\bibfield{author}{\bibinfo{person}{Metashape}.}
  \bibinfo{year}{2020}\natexlab{}.
\newblock \bibinfo{title}{Agisoft Metashape (Version 1.8.4) (Software)}.
\newblock
\newblock
\newblock
\shownote{Retrieved from: \url{https://www.agisoft.com/downloads/installer/}.}


\bibitem[\protect\citeauthoryear{Mihajlovic, Bansal, Zollh{\"{o}}fer, Tang, and
  Saito}{Mihajlovic et~al\mbox{.}}{2022}]%
        {mihajlovic22_keypointnerf}
\bibfield{author}{\bibinfo{person}{Marko Mihajlovic}, \bibinfo{person}{Aayush
  Bansal}, \bibinfo{person}{Michael Zollh{\"{o}}fer}, \bibinfo{person}{Siyu
  Tang}, {and} \bibinfo{person}{Shunsuke Saito}.}
  \bibinfo{year}{2022}\natexlab{}.
\newblock \showarticletitle{KeypointNeRF: Generalizing Image-Based Volumetric
  Avatars Using Relative Spatial Encoding of Keypoints}. In
  \bibinfo{booktitle}{\emph{Eur. Conf. Comput. Vis.}}
  \emph{(\bibinfo{series}{Lecture Notes in Computer Science},
  Vol.~\bibinfo{volume}{13675})}. \bibinfo{publisher}{Springer},
  \bibinfo{pages}{179--197}.
\newblock


\bibitem[\protect\citeauthoryear{Mildenhall, Srinivasan, Tancik, Barron,
  Ramamoorthi, and Ng}{Mildenhall et~al\mbox{.}}{2022}]%
        {mildenhall2022_nerf}
\bibfield{author}{\bibinfo{person}{Ben Mildenhall}, \bibinfo{person}{Pratul~P.
  Srinivasan}, \bibinfo{person}{Matthew Tancik}, \bibinfo{person}{Jonathan~T.
  Barron}, \bibinfo{person}{Ravi Ramamoorthi}, {and} \bibinfo{person}{Ren Ng}.}
  \bibinfo{year}{2022}\natexlab{}.
\newblock \showarticletitle{NeRF: representing scenes as neural radiance fields
  for view synthesis}.
\newblock \bibinfo{journal}{\emph{Commun. {ACM}}} \bibinfo{volume}{65},
  \bibinfo{number}{1} (\bibinfo{year}{2022}), \bibinfo{pages}{99--106}.
\newblock


\bibitem[\protect\citeauthoryear{M{\"{u}}ller, Evans, Schied, and
  Keller}{M{\"{u}}ller et~al\mbox{.}}{2022}]%
        {mueller2022_instant-ngp}
\bibfield{author}{\bibinfo{person}{Thomas M{\"{u}}ller}, \bibinfo{person}{Alex
  Evans}, \bibinfo{person}{Christoph Schied}, {and} \bibinfo{person}{Alexander
  Keller}.} \bibinfo{year}{2022}\natexlab{}.
\newblock \showarticletitle{Instant neural graphics primitives with a
  multiresolution hash encoding}.
\newblock \bibinfo{journal}{\emph{ACM Trans. Graph.}} \bibinfo{volume}{41},
  \bibinfo{number}{4} (\bibinfo{year}{2022}), \bibinfo{pages}{102:1--102:15}.
\newblock


\bibitem[\protect\citeauthoryear{Nagano, Seo, Xing, Wei, Li, Saito, Agarwal,
  Fursund, and Li}{Nagano et~al\mbox{.}}{2018}]%
        {nagano2018_pagan}
\bibfield{author}{\bibinfo{person}{Koki Nagano}, \bibinfo{person}{Jaewoo Seo},
  \bibinfo{person}{Jun Xing}, \bibinfo{person}{Lingyu Wei},
  \bibinfo{person}{Zimo Li}, \bibinfo{person}{Shunsuke Saito},
  \bibinfo{person}{Aviral Agarwal}, \bibinfo{person}{Jens Fursund}, {and}
  \bibinfo{person}{Hao Li}.} \bibinfo{year}{2018}\natexlab{}.
\newblock \showarticletitle{paGAN: real-time avatars using dynamic textures}.
\newblock \bibinfo{journal}{\emph{ACM Trans. Graph.}} \bibinfo{volume}{37},
  \bibinfo{number}{6} (\bibinfo{year}{2018}), \bibinfo{pages}{258}.
\newblock


\bibitem[\protect\citeauthoryear{Or{-}El, Luo, Shan, Shechtman, Park, and
  Kemelmacher{-}Shlizerman}{Or{-}El et~al\mbox{.}}{2022}]%
        {or-el2022_stylesdf}
\bibfield{author}{\bibinfo{person}{Roy Or{-}El}, \bibinfo{person}{Xuan Luo},
  \bibinfo{person}{Mengyi Shan}, \bibinfo{person}{Eli Shechtman},
  \bibinfo{person}{Jeong~Joon Park}, {and} \bibinfo{person}{Ira
  Kemelmacher{-}Shlizerman}.} \bibinfo{year}{2022}\natexlab{}.
\newblock \showarticletitle{StyleSDF: High-Resolution 3D-Consistent Image and
  Geometry Generation}. In \bibinfo{booktitle}{\emph{IEEE Conf. Comput. Vis.
  Pattern Recog.}} \bibinfo{publisher}{{IEEE}}, \bibinfo{pages}{13493--13503}.
\newblock


\bibitem[\protect\citeauthoryear{Park, Florence, Straub, Newcombe, and
  Lovegrove}{Park et~al\mbox{.}}{2019}]%
        {park2019_deepsdf}
\bibfield{author}{\bibinfo{person}{Jeong~Joon Park}, \bibinfo{person}{Peter
  Florence}, \bibinfo{person}{Julian Straub}, \bibinfo{person}{Richard~A.
  Newcombe}, {and} \bibinfo{person}{Steven Lovegrove}.}
  \bibinfo{year}{2019}\natexlab{}.
\newblock \showarticletitle{DeepSDF: Learning Continuous Signed Distance
  Functions for Shape Representation}. In \bibinfo{booktitle}{\emph{IEEE Conf.
  Comput. Vis. Pattern Recog.}} \bibinfo{publisher}{Computer Vision Foundation
  / {IEEE} Computer Society}, \bibinfo{pages}{165--174}.
\newblock


\bibitem[\protect\citeauthoryear{Park, Sinha, Barron, Bouaziz, Goldman, Seitz,
  and Martin{-}Brualla}{Park et~al\mbox{.}}{2021a}]%
        {park2021_nerfies}
\bibfield{author}{\bibinfo{person}{Keunhong Park}, \bibinfo{person}{Utkarsh
  Sinha}, \bibinfo{person}{Jonathan~T. Barron}, \bibinfo{person}{Sofien
  Bouaziz}, \bibinfo{person}{Dan~B. Goldman}, \bibinfo{person}{Steven~M.
  Seitz}, {and} \bibinfo{person}{Ricardo Martin{-}Brualla}.}
  \bibinfo{year}{2021}\natexlab{a}.
\newblock \showarticletitle{Nerfies: Deformable Neural Radiance Fields}. In
  \bibinfo{booktitle}{\emph{Int. Conf. Comput. Vis.}}
  \bibinfo{publisher}{{IEEE}}, \bibinfo{pages}{5845--5854}.
\newblock


\bibitem[\protect\citeauthoryear{Park, Sinha, Hedman, Barron, Bouaziz, Goldman,
  Martin{-}Brualla, and Seitz}{Park et~al\mbox{.}}{2021b}]%
        {park2021_hypernerf}
\bibfield{author}{\bibinfo{person}{Keunhong Park}, \bibinfo{person}{Utkarsh
  Sinha}, \bibinfo{person}{Peter Hedman}, \bibinfo{person}{Jonathan~T. Barron},
  \bibinfo{person}{Sofien Bouaziz}, \bibinfo{person}{Dan~B. Goldman},
  \bibinfo{person}{Ricardo Martin{-}Brualla}, {and} \bibinfo{person}{Steven~M.
  Seitz}.} \bibinfo{year}{2021}\natexlab{b}.
\newblock \showarticletitle{HyperNeRF: a higher-dimensional representation for
  topologically varying neural radiance fields}.
\newblock \bibinfo{journal}{\emph{ACM Trans. Graph.}} \bibinfo{volume}{40},
  \bibinfo{number}{6} (\bibinfo{year}{2021}), \bibinfo{pages}{238:1--238:12}.
\newblock


\bibitem[\protect\citeauthoryear{Parkhi, Vedaldi, and Zisserman}{Parkhi
  et~al\mbox{.}}{2015}]%
        {Parkhi15}
\bibfield{author}{\bibinfo{person}{Omkar~M. Parkhi}, \bibinfo{person}{Andrea
  Vedaldi}, {and} \bibinfo{person}{Andrew Zisserman}.}
  \bibinfo{year}{2015}\natexlab{}.
\newblock \showarticletitle{Deep Face Recognition}. In
  \bibinfo{booktitle}{\emph{British Machine Vision Conference}}.
\newblock


\bibitem[\protect\citeauthoryear{Raj, Zollh{\"{o}}fer, Simon, Saragih, Saito,
  Hays, and Lombardi}{Raj et~al\mbox{.}}{2021}]%
        {raj21_pixel-aligned}
\bibfield{author}{\bibinfo{person}{Amit Raj}, \bibinfo{person}{Michael
  Zollh{\"{o}}fer}, \bibinfo{person}{Tomas Simon}, \bibinfo{person}{Jason~M.
  Saragih}, \bibinfo{person}{Shunsuke Saito}, \bibinfo{person}{James Hays},
  {and} \bibinfo{person}{Stephen Lombardi}.} \bibinfo{year}{2021}\natexlab{}.
\newblock \showarticletitle{Pixel-Aligned Volumetric Avatars}. In
  \bibinfo{booktitle}{\emph{IEEE Conf. Comput. Vis. Pattern Recog.}}
  \bibinfo{publisher}{Computer Vision Foundation / {IEEE} Computer Society},
  \bibinfo{pages}{11733--11742}.
\newblock


\bibitem[\protect\citeauthoryear{Ramon, Triginer, Escur, Pumarola, Giraldez,
  Gir{\'{o}}{-}i{-}Nieto, and Moreno{-}Noguer}{Ramon et~al\mbox{.}}{2021}]%
        {ramon2021_h3dnet}
\bibfield{author}{\bibinfo{person}{Eduard Ramon}, \bibinfo{person}{Gil
  Triginer}, \bibinfo{person}{Janna Escur}, \bibinfo{person}{Albert Pumarola},
  \bibinfo{person}{Jaime~Garcia Giraldez}, \bibinfo{person}{Xavier
  Gir{\'{o}}{-}i{-}Nieto}, {and} \bibinfo{person}{Francesc Moreno{-}Noguer}.}
  \bibinfo{year}{2021}\natexlab{}.
\newblock \showarticletitle{H3D-Net: Few-Shot High-Fidelity 3D Head
  Reconstruction}. In \bibinfo{booktitle}{\emph{Int. Conf. Comput. Vis.}}
  \bibinfo{publisher}{{IEEE}}, \bibinfo{pages}{5600--5609}.
\newblock


\bibitem[\protect\citeauthoryear{Ren, Lattas, Gecer, Deng, Ma, Yang, and
  Zafeiriou}{Ren et~al\mbox{.}}{2022}]%
        {ren2022_facial-geom}
\bibfield{author}{\bibinfo{person}{Xingyu Ren}, \bibinfo{person}{Alexandros
  Lattas}, \bibinfo{person}{Baris Gecer}, \bibinfo{person}{Jiankang Deng},
  \bibinfo{person}{Chao Ma}, \bibinfo{person}{Xiaokang Yang}, {and}
  \bibinfo{person}{Stefanos Zafeiriou}.} \bibinfo{year}{2022}\natexlab{}.
\newblock \showarticletitle{Facial Geometric Detail Recovery via Implicit
  Representation}.
\newblock \bibinfo{journal}{\emph{CoRR}}  \bibinfo{volume}{abs/2203.09692}
  (\bibinfo{year}{2022}).
\newblock


\bibitem[\protect\citeauthoryear{Shamai, Slossberg, and Kimmel}{Shamai
  et~al\mbox{.}}{2019}]%
        {shamai2019_synthesizing-facial}
\bibfield{author}{\bibinfo{person}{Gil Shamai}, \bibinfo{person}{Ron
  Slossberg}, {and} \bibinfo{person}{Ron Kimmel}.}
  \bibinfo{year}{2019}\natexlab{}.
\newblock \showarticletitle{Synthesizing Facial Photometries and Corresponding
  Geometries Using Generative Adversarial Networks}.
\newblock \bibinfo{journal}{\emph{ACM Trans. Multimedia Comput. Commun. Appl.}}
  \bibinfo{volume}{15}, \bibinfo{number}{3s} (\bibinfo{year}{2019}).
\newblock


\bibitem[\protect\citeauthoryear{Sun, Wu, Huang, Zhang, Wang, and Li}{Sun
  et~al\mbox{.}}{2022}]%
        {sun2022_controllable}
\bibfield{author}{\bibinfo{person}{Keqiang Sun}, \bibinfo{person}{Shangzhe Wu},
  \bibinfo{person}{Zhaoyang Huang}, \bibinfo{person}{Ning Zhang},
  \bibinfo{person}{Quan Wang}, {and} \bibinfo{person}{Hongsheng Li}.}
  \bibinfo{year}{2022}\natexlab{}.
\newblock \showarticletitle{Controllable 3D Face Synthesis with Conditional
  Generative Occupancy Fields}.
\newblock \bibinfo{journal}{\emph{CoRR}}  \bibinfo{volume}{abs/2206.08361}
  (\bibinfo{year}{2022}).
\newblock


\bibitem[\protect\citeauthoryear{Takikawa, Litalien, Yin, Kreis, Loop,
  Nowrouzezahrai, Jacobson, McGuire, and Fidler}{Takikawa
  et~al\mbox{.}}{2021}]%
        {takikawa2021_nglod}
\bibfield{author}{\bibinfo{person}{Towaki Takikawa}, \bibinfo{person}{Joey
  Litalien}, \bibinfo{person}{Kangxue Yin}, \bibinfo{person}{Karsten Kreis},
  \bibinfo{person}{Charles~T. Loop}, \bibinfo{person}{Derek Nowrouzezahrai},
  \bibinfo{person}{Alec Jacobson}, \bibinfo{person}{Morgan McGuire}, {and}
  \bibinfo{person}{Sanja Fidler}.} \bibinfo{year}{2021}\natexlab{}.
\newblock \showarticletitle{Neural Geometric Level of Detail: Real-Time
  Rendering With Implicit 3D Shapes}. In \bibinfo{booktitle}{\emph{IEEE Conf.
  Comput. Vis. Pattern Recog.}} \bibinfo{publisher}{Computer Vision Foundation
  / {IEEE} Computer Society}, \bibinfo{pages}{11358--11367}.
\newblock


\bibitem[\protect\citeauthoryear{Tang}{Tang}{2022}]%
        {tang2022_torch-ngp}
\bibfield{author}{\bibinfo{person}{Jiaxiang Tang}.}
  \bibinfo{year}{2022}\natexlab{}.
\newblock \bibinfo{title}{Torch-ngp: a PyTorch implementation of instant-ngp}.
\newblock
\newblock
\newblock
\shownote{https://github.com/ashawkey/torch-ngp.}


\bibitem[\protect\citeauthoryear{Tewari, Elgharib, R., Bernard, Seidel,
  P{\'{e}}rez, Zollh{\"{o}}fer, and Theobalt}{Tewari et~al\mbox{.}}{2020}]%
        {tewari2020_pie}
\bibfield{author}{\bibinfo{person}{Ayush Tewari}, \bibinfo{person}{Mohamed
  Elgharib}, \bibinfo{person}{Mallikarjun~B. R.}, \bibinfo{person}{Florian
  Bernard}, \bibinfo{person}{Hans{-}Peter Seidel}, \bibinfo{person}{Patrick
  P{\'{e}}rez}, \bibinfo{person}{Michael Zollh{\"{o}}fer}, {and}
  \bibinfo{person}{Christian Theobalt}.} \bibinfo{year}{2020}\natexlab{}.
\newblock \showarticletitle{{PIE:} portrait image embedding for semantic
  control}.
\newblock \bibinfo{journal}{\emph{ACM Trans. Graph.}} \bibinfo{volume}{39},
  \bibinfo{number}{6} (\bibinfo{year}{2020}), \bibinfo{pages}{223:1--223:14}.
\newblock


\bibitem[\protect\citeauthoryear{Tewari, Thies, Mildenhall, Srinivasan,
  Tretschk, Wang, Lassner, Sitzmann, Martin{-}Brualla, Lombardi, Simon,
  Theobalt, Nie{\ss}ner, Barron, Wetzstein, Zollh{\"{o}}fer, and
  Golyanik}{Tewari et~al\mbox{.}}{2022}]%
        {tewari2022_advances-neural-render}
\bibfield{author}{\bibinfo{person}{Ayush Tewari}, \bibinfo{person}{Justus
  Thies}, \bibinfo{person}{Ben Mildenhall}, \bibinfo{person}{Pratul~P.
  Srinivasan}, \bibinfo{person}{Edgar Tretschk}, \bibinfo{person}{Yifan Wang},
  \bibinfo{person}{Christoph Lassner}, \bibinfo{person}{Vincent Sitzmann},
  \bibinfo{person}{Ricardo Martin{-}Brualla}, \bibinfo{person}{Stephen
  Lombardi}, \bibinfo{person}{Tomas Simon}, \bibinfo{person}{Christian
  Theobalt}, \bibinfo{person}{Matthias Nie{\ss}ner},
  \bibinfo{person}{Jonathan~T. Barron}, \bibinfo{person}{Gordon Wetzstein},
  \bibinfo{person}{Michael Zollh{\"{o}}fer}, {and} \bibinfo{person}{Vladislav
  Golyanik}.} \bibinfo{year}{2022}\natexlab{}.
\newblock \showarticletitle{Advances in Neural Rendering}.
\newblock \bibinfo{journal}{\emph{Comput. Graph. Forum}} \bibinfo{volume}{41},
  \bibinfo{number}{2} (\bibinfo{year}{2022}), \bibinfo{pages}{703--735}.
\newblock


\bibitem[\protect\citeauthoryear{Tewari, Zollh{\"{o}}fer, Garrido, Bernard,
  Kim, P{\'{e}}rez, and Theobalt}{Tewari et~al\mbox{.}}{2018}]%
        {tewari2018_self-supervised}
\bibfield{author}{\bibinfo{person}{Ayush Tewari}, \bibinfo{person}{Michael
  Zollh{\"{o}}fer}, \bibinfo{person}{Pablo Garrido}, \bibinfo{person}{Florian
  Bernard}, \bibinfo{person}{Hyeongwoo Kim}, \bibinfo{person}{Patrick
  P{\'{e}}rez}, {and} \bibinfo{person}{Christian Theobalt}.}
  \bibinfo{year}{2018}\natexlab{}.
\newblock \showarticletitle{Self-Supervised Multi-Level Face Model Learning for
  Monocular Reconstruction at Over 250 Hz}. In \bibinfo{booktitle}{\emph{IEEE
  Conf. Comput. Vis. Pattern Recog.}} \bibinfo{publisher}{Computer Vision
  Foundation / {IEEE} Computer Society}, \bibinfo{pages}{2549--2559}.
\newblock


\bibitem[\protect\citeauthoryear{Thies, Zollh{\"{o}}fer, and Nie{\ss}ner}{Thies
  et~al\mbox{.}}{2019a}]%
        {thies2019_dnr}
\bibfield{author}{\bibinfo{person}{Justus Thies}, \bibinfo{person}{Michael
  Zollh{\"{o}}fer}, {and} \bibinfo{person}{Matthias Nie{\ss}ner}.}
  \bibinfo{year}{2019}\natexlab{a}.
\newblock \showarticletitle{Deferred neural rendering: image synthesis using
  neural textures}.
\newblock \bibinfo{journal}{\emph{ACM Trans. Graph.}} \bibinfo{volume}{38},
  \bibinfo{number}{4} (\bibinfo{year}{2019}), \bibinfo{pages}{66:1--66:12}.
\newblock


\bibitem[\protect\citeauthoryear{Thies, Zollh{\"o}fer, Stamminger, Theobalt,
  and Nie{\ss}ner}{Thies et~al\mbox{.}}{2016}]%
        {thies2016face}
\bibfield{author}{\bibinfo{person}{J. Thies}, \bibinfo{person}{M.
  Zollh{\"o}fer}, \bibinfo{person}{M. Stamminger}, \bibinfo{person}{C.
  Theobalt}, {and} \bibinfo{person}{M. Nie{\ss}ner}.}
  \bibinfo{year}{2016}\natexlab{}.
\newblock \showarticletitle{Face2Face: Real-time Face Capture and Reenactment
  of RGB Videos}. In \bibinfo{booktitle}{\emph{Proc. Computer Vision and
  Pattern Recognition (CVPR), IEEE}}.
\newblock


\bibitem[\protect\citeauthoryear{Thies, Zollh{\"{o}}fer, Stamminger, Theobalt,
  and Nie{\ss}ner}{Thies et~al\mbox{.}}{2019b}]%
        {thies2019_face2face}
\bibfield{author}{\bibinfo{person}{Justus Thies}, \bibinfo{person}{Michael
  Zollh{\"{o}}fer}, \bibinfo{person}{Marc Stamminger},
  \bibinfo{person}{Christian Theobalt}, {and} \bibinfo{person}{Matthias
  Nie{\ss}ner}.} \bibinfo{year}{2019}\natexlab{b}.
\newblock \showarticletitle{Face2Face: real-time face capture and reenactment
  of {RGB} videos}.
\newblock \bibinfo{journal}{\emph{Commun. {ACM}}} \bibinfo{volume}{62},
  \bibinfo{number}{1} (\bibinfo{year}{2019}), \bibinfo{pages}{96--104}.
\newblock


\bibitem[\protect\citeauthoryear{Tran, Liu, and Liu}{Tran
  et~al\mbox{.}}{2019}]%
        {tran2019_towards}
\bibfield{author}{\bibinfo{person}{Luan Tran}, \bibinfo{person}{Feng Liu},
  {and} \bibinfo{person}{Xiaoming Liu}.} \bibinfo{year}{2019}\natexlab{}.
\newblock \showarticletitle{Towards High-Fidelity Nonlinear 3D Face Morphable
  Model}. In \bibinfo{booktitle}{\emph{IEEE Conf. Comput. Vis. Pattern Recog.}}
  \bibinfo{publisher}{Computer Vision Foundation / {IEEE} Computer Society},
  \bibinfo{pages}{1126--1135}.
\newblock


\bibitem[\protect\citeauthoryear{Wang, Chandran, Zoss, Bradley, and
  Gotardo}{Wang et~al\mbox{.}}{2022}]%
        {wang2022_morf}
\bibfield{author}{\bibinfo{person}{Daoye Wang}, \bibinfo{person}{Prashanth
  Chandran}, \bibinfo{person}{Gaspard Zoss}, \bibinfo{person}{Derek Bradley},
  {and} \bibinfo{person}{Paulo F.~U. Gotardo}.}
  \bibinfo{year}{2022}\natexlab{}.
\newblock \showarticletitle{MoRF: Morphable Radiance Fields for Multiview
  Neural Head Modeling}. In \bibinfo{booktitle}{\emph{{SIGGRAPH}}}.
  \bibinfo{publisher}{{ACM}}, \bibinfo{pages}{55:1--55:9}.
\newblock


\bibitem[\protect\citeauthoryear{Wang, Mallya, and Liu}{Wang
  et~al\mbox{.}}{2021b}]%
        {wang2021_one-shot}
\bibfield{author}{\bibinfo{person}{Ting{-}Chun Wang}, \bibinfo{person}{Arun
  Mallya}, {and} \bibinfo{person}{Ming{-}Yu Liu}.}
  \bibinfo{year}{2021}\natexlab{b}.
\newblock \showarticletitle{One-Shot Free-View Neural Talking-Head Synthesis
  for Video Conferencing}. In \bibinfo{booktitle}{\emph{IEEE Conf. Comput. Vis.
  Pattern Recog.}} \bibinfo{publisher}{Computer Vision Foundation / {IEEE}
  Computer Society}, \bibinfo{pages}{10039--10049}.
\newblock


\bibitem[\protect\citeauthoryear{Wang, Bagautdinov, Lombardi, Simon, Saragih,
  Hodgins, and Zollh{\"{o}}fer}{Wang et~al\mbox{.}}{2021a}]%
        {wang2021_learning-crf}
\bibfield{author}{\bibinfo{person}{Ziyan Wang}, \bibinfo{person}{Timur~M.
  Bagautdinov}, \bibinfo{person}{Stephen Lombardi}, \bibinfo{person}{Tomas
  Simon}, \bibinfo{person}{Jason~M. Saragih}, \bibinfo{person}{Jessica~K.
  Hodgins}, {and} \bibinfo{person}{Michael Zollh{\"{o}}fer}.}
  \bibinfo{year}{2021}\natexlab{a}.
\newblock \showarticletitle{Learning Compositional Radiance Fields of Dynamic
  Human Heads}. In \bibinfo{booktitle}{\emph{IEEE Conf. Comput. Vis. Pattern
  Recog.}} \bibinfo{publisher}{Computer Vision Foundation / {IEEE} Computer
  Society}, \bibinfo{pages}{5704--5713}.
\newblock


\bibitem[\protect\citeauthoryear{Wang, Bovik, Sheikh, and Simoncelli}{Wang
  et~al\mbox{.}}{2004}]%
        {wang2004_image-quality}
\bibfield{author}{\bibinfo{person}{Zhou Wang}, \bibinfo{person}{Alan~C. Bovik},
  \bibinfo{person}{Hamid~R. Sheikh}, {and} \bibinfo{person}{Eero~P.
  Simoncelli}.} \bibinfo{year}{2004}\natexlab{}.
\newblock \showarticletitle{Image quality assessment: from error visibility to
  structural similarity}.
\newblock \bibinfo{journal}{\emph{{IEEE} Trans. Image Process.}}
  \bibinfo{volume}{13}, \bibinfo{number}{4} (\bibinfo{year}{2004}),
  \bibinfo{pages}{600--612}.
\newblock


\bibitem[\protect\citeauthoryear{Xiang, Yang, Deng, and Tong}{Xiang
  et~al\mbox{.}}{2022}]%
        {Xiang22GRAM-HD}
\bibfield{author}{\bibinfo{person}{Jianfeng Xiang}, \bibinfo{person}{Jiaolong
  Yang}, \bibinfo{person}{Yu Deng}, {and} \bibinfo{person}{Xin Tong}.}
  \bibinfo{year}{2022}\natexlab{}.
\newblock \bibinfo{title}{GRAM-HD: 3D-Consistent Image Generation at High
  Resolution with Generative Radiance Manifolds}.
\newblock
\newblock
\urldef\tempurl%
\url{https://doi.org/10.48550/ARXIV.2206.07255}
\showDOI{\tempurl}


\bibitem[\protect\citeauthoryear{Yamaguchi, Saito, Nagano, Zhao, Chen,
  Olszewski, Morishima, and Li}{Yamaguchi et~al\mbox{.}}{2018}]%
        {yamaguchi2018_high-fidelity}
\bibfield{author}{\bibinfo{person}{Shugo Yamaguchi}, \bibinfo{person}{Shunsuke
  Saito}, \bibinfo{person}{Koki Nagano}, \bibinfo{person}{Yajie Zhao},
  \bibinfo{person}{Weikai Chen}, \bibinfo{person}{Kyle Olszewski},
  \bibinfo{person}{Shigeo Morishima}, {and} \bibinfo{person}{Hao Li}.}
  \bibinfo{year}{2018}\natexlab{}.
\newblock \showarticletitle{High-fidelity facial reflectance and geometry
  inference from an unconstrained image}.
\newblock \bibinfo{journal}{\emph{ACM Trans. Graph.}} \bibinfo{volume}{37},
  \bibinfo{number}{4} (\bibinfo{year}{2018}), \bibinfo{pages}{162}.
\newblock


\bibitem[\protect\citeauthoryear{Yu, Li, Tancik, Li, Ng, and Kanazawa}{Yu
  et~al\mbox{.}}{2021}]%
        {yu2021_plenoctrees}
\bibfield{author}{\bibinfo{person}{Alex Yu}, \bibinfo{person}{Ruilong Li},
  \bibinfo{person}{Matthew Tancik}, \bibinfo{person}{Hao Li},
  \bibinfo{person}{Ren Ng}, {and} \bibinfo{person}{Angjoo Kanazawa}.}
  \bibinfo{year}{2021}\natexlab{}.
\newblock \showarticletitle{PlenOctrees for Real-time Rendering of Neural
  Radiance Fields}. In \bibinfo{booktitle}{\emph{Int. Conf. Comput. Vis.}}
  \bibinfo{publisher}{{IEEE}}, \bibinfo{pages}{5732--5741}.
\newblock


\bibitem[\protect\citeauthoryear{Zhang, Dai, Tai, and Tang}{Zhang
  et~al\mbox{.}}{2022}]%
        {zhang2022_flnerf}
\bibfield{author}{\bibinfo{person}{Hao Zhang}, \bibinfo{person}{Tianyuan Dai},
  \bibinfo{person}{Yu{-}Wing Tai}, {and} \bibinfo{person}{Chi{-}Keung Tang}.}
  \bibinfo{year}{2022}\natexlab{}.
\newblock \showarticletitle{FLNeRF: 3D Facial Landmarks Estimation in Neural
  Radiance Fields}.
\newblock \bibinfo{journal}{\emph{CoRR}}  \bibinfo{volume}{abs/2211.11202}
  (\bibinfo{year}{2022}).
\newblock


\bibitem[\protect\citeauthoryear{Zhang, Isola, Efros, Shechtman, and
  Wang}{Zhang et~al\mbox{.}}{2018}]%
        {zhang2018_unreasonable-effectiveness}
\bibfield{author}{\bibinfo{person}{Richard Zhang}, \bibinfo{person}{Phillip
  Isola}, \bibinfo{person}{Alexei~A. Efros}, \bibinfo{person}{Eli Shechtman},
  {and} \bibinfo{person}{Oliver Wang}.} \bibinfo{year}{2018}\natexlab{}.
\newblock \showarticletitle{The Unreasonable Effectiveness of Deep Features as
  a Perceptual Metric}. In \bibinfo{booktitle}{\emph{IEEE Conf. Comput. Vis.
  Pattern Recog.}} \bibinfo{publisher}{Computer Vision Foundation / {IEEE}
  Computer Society}, \bibinfo{pages}{586--595}.
\newblock


\bibitem[\protect\citeauthoryear{Zheng, Abrevaya, B{\"{u}}hler, Chen, Black,
  and Hilliges}{Zheng et~al\mbox{.}}{2022}]%
        {zheng2022imavatar}
\bibfield{author}{\bibinfo{person}{Yufeng Zheng},
  \bibinfo{person}{Victoria~Fern{\'{a}}ndez Abrevaya},
  \bibinfo{person}{Marcel~C. B{\"{u}}hler}, \bibinfo{person}{Xu Chen},
  \bibinfo{person}{Michael~J. Black}, {and} \bibinfo{person}{Otmar Hilliges}.}
  \bibinfo{year}{2022}\natexlab{}.
\newblock \showarticletitle{I {M} Avatar: Implicit Morphable Head Avatars from
  Videos}. In \bibinfo{booktitle}{\emph{IEEE Conf. Comput. Vis. Pattern
  Recog.}} \bibinfo{publisher}{{IEEE}}, \bibinfo{pages}{13535--13545}.
\newblock


\bibitem[\protect\citeauthoryear{Zollh{\"{o}}fer, Thies, Garrido, Bradley,
  Beeler, P{\'{e}}rez, Stamminger, Nie{\ss}ner, and Theobalt}{Zollh{\"{o}}fer
  et~al\mbox{.}}{2018}]%
        {zollhofer18_sota-mon-3d}
\bibfield{author}{\bibinfo{person}{Michael Zollh{\"{o}}fer},
  \bibinfo{person}{Justus Thies}, \bibinfo{person}{Pablo Garrido},
  \bibinfo{person}{Derek Bradley}, \bibinfo{person}{Thabo Beeler},
  \bibinfo{person}{Patrick P{\'{e}}rez}, \bibinfo{person}{Marc Stamminger},
  \bibinfo{person}{Matthias Nie{\ss}ner}, {and} \bibinfo{person}{Christian
  Theobalt}.} \bibinfo{year}{2018}\natexlab{}.
\newblock \showarticletitle{State of the Art on Monocular 3D Face
  Reconstruction, Tracking, and Applications}.
\newblock \bibinfo{journal}{\emph{Comput. Graph. Forum}} \bibinfo{volume}{37},
  \bibinfo{number}{2} (\bibinfo{year}{2018}), \bibinfo{pages}{523--550}.
\newblock


\end{thebibliography}

\end{document}